\documentclass[lettersize,journal,compsoc]{IEEEtran}
\usepackage{amsmath,amsfonts}
\usepackage{algorithmic}
\usepackage{array}
\usepackage[caption=false,font=normalsize,labelfont=sf,textfont=sf]{subfig}
\usepackage{textcomp}
\usepackage{stfloats}
\usepackage{url}
\usepackage{verbatim}
\usepackage{graphicx}

\usepackage{xcolor}
\usepackage{amsmath}
\usepackage{bbm}
\usepackage{bm}

\usepackage{algorithm}
\usepackage{algorithmic}

\usepackage{bbding}
\usepackage{makecell}
\usepackage{booktabs}
\usepackage{multicol}
\usepackage{multirow}

\usepackage{caption}

\usepackage{soul}
\usepackage{doi}
\usepackage{mathtools}

\usepackage{hyperref}

\usepackage{orcidlink}

\usepackage{etoolbox}
\usepackage{ragged2e}

\usepackage{multicol}
\usepackage{multirow}

\usepackage[capitalize]{cleveref}
\crefname{section}{Sec.}{Secs.}
\Crefname{section}{Section}{Sections}
\Crefname{table}{Table}{Tables}
\crefname{table}{Tab.}{Tabs.}
\crefname{algorithm}{Algo.}{Algos.}

\usepackage{xspace}
\makeatletter
\DeclareRobustCommand\onedot{\futurelet\@let@token\@onedot}
\def\@onedot{\ifx\@let@token.\else.\null\fi\xspace}
\def\eg{\emph{e.g}\onedot} 
\def\ie{\emph{i.e}\onedot} 
 
\def\etc{\emph{etc}\onedot} \def\vs{\emph{vs}\onedot}
\def\wrt{w.r.t\onedot} 

\def\aka{\emph{a.k.a}\onedot}

\def\BibTeX{{\rm B\kern-.05em{\sc i\kern-.025em b}\kern-.08em
    T\kern-.1667em\lower.7ex\hbox{E}\kern-.125emX}}
\usepackage{balance}

\begin{document}

\title{Ultra-High-Resolution Image Synthesis: Data, Method and Evaluation}

\author{Jinjin Zhang$^{\orcidlink{0009-0007-8419-148X}}$, Qiuyu Huang$^{\orcidlink{0009-0007-5047-4094}}$, Junjie Liu$^{\orcidlink{0000-0001-7373-9074}}$, Xiefan Guo$^{\orcidlink{0009-0002-5936-3896}}$ and Di Huang$^{\orcidlink{0000-0002-2412-9330}}$,~\IEEEmembership{Senior Member,~IEEE}
\IEEEcompsocitemizethanks{
\IEEEcompsocthanksitem J. Zhang, Q. Huang, X. Guo and D. Huang are with the State Key Laboratory of Complex and Critical Software Environment, School of Computer Science and Engineering, Beihang University, Beijing 100191, China 
(email: \{jinjin.zhang, huangqiuyu, xfguo, dhuang\}@buaa.edu.cn ).

\IEEEcompsocthanksitem Corresponding author: Di Huang.

\IEEEcompsocthanksitem J. Liu is with Meituan, Beijing 100102, China (email: liujunjie10@meituan.com).
}
}


\IEEEtitleabstractindextext{
\begin{abstract}
\justifying
Ultra-high-resolution image synthesis holds significant potential, yet remains an underexplored challenge due to the absence of standardized benchmarks and computational constraints.
In this paper, we establish Aesthetic-4K, a meticulously curated dataset containing dedicated training and evaluation subsets specifically designed for comprehensive research on ultra-high-resolution image synthesis.
This dataset consists of high-quality 4K images accompanied by descriptive captions generated by GPT-4o. 
Furthermore, we propose Diffusion-4K, an innovative framework for the direct generation of ultra-high-resolution images.
Our approach incorporates the Scale Consistent Variational Auto-Encoder (SC-VAE) and Wavelet-based Latent Fine-tuning (WLF), which are designed for efficient visual token compression and the capture of intricate details in ultra-high-resolution images, thereby facilitating direct training with photorealistic 4K data.
This method is applicable to various latent diffusion models and demonstrates its efficacy in synthesizing highly detailed 4K images.
Additionally, we propose novel metrics, namely the GLCM Score and Compression Ratio, to assess the texture richness and fine details in local patches, in conjunction with holistic measures such as FID, Aesthetics, and CLIPScore, enabling a thorough and multifaceted evaluation of ultra-high-resolution image synthesis.
Consequently, Diffusion-4K achieves impressive performance in ultra-high-resolution image synthesis, particularly when powered by state-of-the-art large-scale diffusion models (\eg, Flux-12B). 
The source code is publicly available at \url{https://github.com/zhang0jhon/diffusion-4k}.
\end{abstract}

}

\maketitle

\IEEEdisplaynontitleabstractindextext

\IEEEpeerreviewmaketitle

\section{Introduction}
\IEEEPARstart{D}{iffusion} models have demonstrated remarkable efficacy in modeling high-dimensional, perceptual data, such as images~\cite{ho2020denoising, song2020denoising, song2020score, nichol2021improved, song2021maximum, vahdat2021score, dhariwal2021diffusion, karras2022elucidating, nichol2022glide, ho2022classifier, zhan2023multimodal, yang2023diffusion, peebles2023scalable, croitoru2023diffusion, xia2024dmt}.
These models have significantly propelled advancements in deep generative modeling, particularly with prominent implementations such as Imagen~\cite{saharia2022photorealistic, baldridge2024imagen}, DALL·E 2/3~\cite{ramesh2022hierarchical, betker2023improving} and Stable Diffusion~\cite{rombach2022high}, \etc.
In recent years, latent diffusion models have made substantial strides in text-to-image synthesis, showcasing impressive generalization capabilities, especially at high resolutions~\cite{podell2023sdxl, esser2024scaling, chen2024pixart, li2024playground, liu2024playground, zhang2025diffusion4k}. 
Notably, the adoption of transformer architectures in place of convolutional U-Nets has yielded promising results, particularly as model scalability increases. Examples of such advancements include Stable Diffusion 3 (SD3) with 8B parameters~\cite{esser2024scaling}, Flux with 12B parameters~\cite{Flux:2024:Online}, and Playground v3 with 24B parameters~\cite{liu2024playground}.
On another front, flow-based models~\cite{lipman2022flow, liu2022flow, albergo2022building}, which utilize data or velocity prediction, have emerged as a competitive alternative, offering faster convergence and improved performance~\cite{karras2022elucidating, albergo2022building, ma2024sit, xie2024sana}.

Despite significant advancements, most latent diffusion models primarily focus on training and generating images at $1024 \times 1024$ resolution, leaving the direct synthesis of ultra-high-resolution images largely underexplored.
Direct training and generation of 4K images (typically referring to a resolution of approximately 4096 pixels) hold significant value in practical applications, such as industrial manufacturing, film production, and game development, \etc.
However, this task necessitates substantial computational resources, particularly as model parameters continue to increase.
Recent approaches, including PixArt-$\Sigma$~\cite{chen2024pixart} and Sana~\cite{xie2024sana, xie2025sana}, have addressed the challenge of direct ultra-high-resolution image synthesis at 4K resolution using private high-quality datasets, showcasing the potential of scalable latent diffusion transformer architectures, utilizing techniques such as token compression or linear attention mechanisms.
Both PixArt-$\Sigma$ with 0.6B parameters~\cite{chen2024pixart} and Sana with 1.6B/4.8B parameters~\cite{xie2024sana, xie2025sana} are primarily designed to prioritize the efficiency of ultra-high-resolution image generation, however, \textbf{the intrinsic benefits of 4K images, such as capturing high-frequency details and rich textures, are overlooked within their optimization frameworks}.
Furthermore, these approaches \textbf{lack comprehensive assessments} for ultra-high-resolution image synthesis due to the absence of standardized benchmarks, thus impeding further progress in this critical area of research.

In this paper, we introduce Aesthetic-4K, a high-quality dataset comprising curated training and evaluation sets of ultra-high-resolution images, accompanied by corresponding captions generated by GPT-4o~\cite{hurst2024gpt}.
Furthermore, we propose Diffusion-4K, a novel framework for the direct synthesis of ultra-high-resolution images, designed to be compatible with various latent diffusion models.
Specifically, we design the Scale Consistent Variational Auto-Encoder (SC-VAE), which efficiently compresses visual tokens while maintaining consistency across multi-scale feature maps, thereby significantly reducing the memory and computational overhead.
In parallel, we propose Wavelet-based Latent Fine-tuning (WLF) to enhance high-frequency components while preserving low-frequency approximations in the synthesis of ultra-high-resolution images.
Moreover, most existing evaluation metrics, such as Fréchet Inception Distance (FID)~\cite{heusel2017gans}, Aesthetics~\cite{schuhmann2022laion} and CLIPScore~\cite{hessel2021clipscore}, primarily provide holistic measures at lower resolutions, which are inadequate for the comprehensive benchmarking in ultra-high-resolution image synthesis.
To address these limitations, we propose new metrics, Gray Level Co-occurrence Matrix (GLCM) Score and Compression Ratio, focusing on the assessment of rich textures and fine details in local patches, an area that has yet to be explored, aiming to establish a comprehensive assessment for ultra-high-resolution image synthesis.
We conduct experiments with state-of-the-art latent diffusion models, including SD3-2B~\cite{esser2024scaling} and Flux-12B~\cite{Flux:2024:Online}, to demonstrate the advantages of our approach in synthesizing highly detailed 4K images.
Consequently, our method achieves superior performance in ultra-high-resolution image synthesis on the Aesthetic-4K dataset, highlighting the effectiveness of the proposed framework.

The main contributions are summarized as follows:
\begin{itemize}
\item We construct Aesthetic-4K, a high-quality dataset comprising  standardized training and evaluation sets for ultra-high-resolution image synthesis, characterized by exceptional visual quality and fine details.
\item We propose Diffusion-4K, which integrates SC-VAE and WLF, compatible with various latent diffusion models, emphasizing the generation of ultra-high-resolution images with fine details.
\item We design novel indicators for image quality assessment at the local patch level, which exhibit strong alignment with human perceptual cognition and, when combined with existing holistic metrics, enable a comprehensive and multifaceted evaluation of ultra-high-resolution image generation.
\item Extensive experimental results demonstrate the effectiveness and generalization of our proposed method in 4K image synthesis, particularly when applied to state-of-the-art large-scale diffusion transformers. 
\end{itemize}

A preliminary version of this study was previously published in~\cite{zhang2025diffusion4k}. 
This paper introduces significant improvements in the following aspects:
(i) \textbf{Scale Consistent Variational Auto-Encoder (SC-VAE)}: We propose SC-VAE, a novel and efficient VAE for visual token compression, as detailed in \cref{sec:sc_vae}, and fine-tune it on the large-scale Segment Anything 1 Billion (SA-1B) dataset~\cite{kirillov2023segment}.
Quantitative and qualitative evaluations on the Aesthetic-4K dataset demonstrate significant improvements in both ultra-high-resolution image reconstruction and generation tasks compared to the previously proposed partitioned VAE~\cite{zhang2025diffusion4k}.
Furthermore, we conduct an ablation study on the scale consistency mechanism of SC-VAE, as shown in \cref{tab:ablation_vae_reconstruction}, demonstrating its superiority over vanilla VAE fine-tuning methods~\cite{esser2021taming, rombach2022high}.
(ii) \textbf{Enhanced Training Dataset for Scalability Analysis}: We introduce Aesthetic-Train-V2, a significantly expanded training set for scalability analysis that consists of 105,288 high-quality image-text pairs, representing a nearly 9-fold increase over the 12,015 pairs in Aesthetic-Train~\cite{zhang2025diffusion4k}.
Furthermore, we validate the effectiveness and generalization of our approach through quantitative and qualitative scalability experiments in \cref{sec:scalability_analysis}, particularly demonstrating improvements in the fine details of ultra-high-resolution images with scalable high-quality data.
(iii) \textbf{Comprehensive Evaluation Against State-of-the-Art Models}: We conduct both quantitative and qualitative evaluations on Aesthetic-Eval in \cref{sec:experimental_results}, comparing our method against state-of-the-art latent diffusion models for direct ultra-high-resolution image synthesis. These include PixArt-$\Sigma$~\cite{chen2024pixart}, which utilizes token compression, and Sana~\cite{xie2024sana}, which employs a linear diffusion transformer. 
Results demonstrate that our approach achieves superior performance in generating structured textures and intricate fine details.
(iv) \textbf{Human and AI Preference Studies}: We additionally  present qualitative results and conduct both human and AI-based preference studies in comparison to existing ultra-high-resolution image synthesis approaches. 
As illustrated in~\cref{fig:win_rate}, our approach achieves consistent improvements across multiple dimensions, including visual aesthetics, prompt adherence, and detail fidelity, when compared with our previous work~\cite{zhang2025diffusion4k}. 
Moreover, our method obtains higher human preference scores relative to state-of-the-art models, including both PixArt-$\Sigma$~\cite{chen2024pixart} and Sana~\cite{xie2024sana}.
\section{Related work}

\subsection{Latent Diffusion Models} 

Stable Diffusion (SD)~\cite{rombach2022high} introduces latent diffusion models, which performs the diffusion process in compressed latent space using Variational Auto-Encoder (VAE)~\cite{kingma2013auto, van2017neural}.
Widely adopted VAEs~\cite{peebles2023scalable, esser2024scaling, chen2023pixart} in latent diffusion models typically employ a down-sampling factor of $F=8$, compressing pixel space $\mathbb{R}^{H \times W \times 3}$ into latent space $\mathbb{R}^{\frac{H}{F} \times \frac{W}{F} \times C}$, where $H$ and $W$ represent height and width, respectively, and $C$ denotes the channel of the latent space.
In recent developments within latent diffusion models, the Diffusion Transformer (DiT)~\cite{peebles2023scalable} has made significant progress by replacing the conventional U-Net backbone with a transformer architecture that operates on latent patches.
Typically, the patch size of DiT is set to $P=2$, resulting in $\frac{H}{FP} \times \frac{H}{FP}$ tokens.
The transformer architecture exhibits excellent scalability in latent diffusion models, as evidenced by state-of-the-art models such as DALL·E 2/3~\cite{ramesh2022hierarchical, betker2023improving}, DiffiT~\cite{hatamizadeh2024diffit}, PixArt~\cite{chen2023pixart, chen2024pixart}, SD3~\cite{esser2024scaling}, Flux~\cite{Flux:2024:Online}, and Playground~\cite{li2024playground, liu2024playground}.
Specifically, SD3~\cite{esser2024scaling} and Flux~\cite{Flux:2024:Online} incorporate an enhanced MM-DiT architecture for latent diffusion models, designed to handle different domains, here text and image tokens, using different sets of trainable model weights. 
Notably, Flux further enhances the Sinusoidal Positional Encoding (PE)~\cite{vaswani2017attention} used in SD3 by incorporating the Rotary Position Embedding (RoPE)~\cite{su2024roformer}.
The proposed multi-modal diffusion backbone, MM-DiT, significantly  improves modality-specific representations, demonstrating a marked performance boost over both the cross-attention and vanilla variants in DiTs.

In text-to-image synthesis, the text encoder plays a crucial role in ensuring prompt coherence.
DALL·E 3~\cite{betker2023improving} demonstrates that training with descriptive image captions can significantly enhance prompt coherence in text-to-image diffusion models.
SD employs the pretrained CLIP~\cite{radford2021learning} as its text encoder but is constrained by the limited 77 text tokens.
In contrast, subsequent diffusion models, such as Imagen~\cite{saharia2022photorealistic} and PixArt~\cite{chen2023pixart, chen2024pixart}, utilize T5-XXL~\cite{raffel2020exploring} with 4.7B parameters for text feature extraction to address the token limitation.
Recent advancements, such as SD3~\cite{esser2024scaling} and Flux~\cite{Flux:2024:Online}, integrate both CLIP and T5-XXL for improved text understanding. 
Furthermore, Sana~\cite{xie2024sana} employs the latest efficient decoder-only Large Language Model (LLM), Gemma 2~\cite{team2024gemma} with 2B parameters, as its text encoder to enhance both understanding and reasoning capabilities related to text prompts.

\subsection{High-Resolution Image Synthesis}

High-resolution image generation is of significant value across various practical applications, including industry and entertainment.
Generative Adversarial Networks (GANs)~\cite{goodfellow2014generative, karras2017progressive, brock2018large, karras2019style, karras2020analyzing, karras2021alias, chai2022any, kang2023scaling} have long been a dominant family of generative models for natural image synthesis, demonstrating impressive capabilities, particularly in single-category domains. 
Autoregressive (AR) models, such as VQ-VAE~\cite{van2017neural, razavi2019generating}, VQ-GAN~\cite{esser2021taming}, DALL·E~\cite{ramesh2021zero}, Muse~\cite{chang2023muse}, Parti~\cite{yu2022scaling}, VAR~\cite{tian2024visual}, MAR~\cite{li2024autoregressive}, have also witnessed rapid growth in image generation.
Specifically, VQ-GAN~\cite{esser2021taming} learns an effective codebook of context-rich visual constituents and their global compositions using latent transformers, enabling the synthesis of high-resolution images.
GigaGAN~\cite{kang2023scaling} reintroduces multi-scale training and achieves stable and scalable GAN training on large-scale datasets, facilitating the synthesis of ultra-high-resolution images.

In the case of state-of-the-art latent diffusion models~\cite{rombach2022high, ramesh2021zero, ramesh2022hierarchical, podell2023sdxl, sauer2024adversarial, sauer2024fast}, current advancements are typically trained to synthesize images at $1024 \times 1024$ resolution, primarily due to the computational complexity constraints. 
Notably, increasing image resolution results in quadratic computational costs, posing significant challenges for 4K image synthesis.
Several training-free fusion approaches for 4K image generation have been proposed, leveraging existing latent diffusion models~\cite{bar2023multidiffusion, du2024demofusion, haji2024elasticdiffusion}. 
Additionally, Stable Cascade~\cite{pernias2023wurstchen} employs multiple diffusion networks to increase resolution progressively. 
However, these ensemble approaches can introduce cumulative errors, which may degrade image quality.
PixArt-$\Sigma$~\cite{chen2024pixart} pioneers direct image generation close to 4K resolution ($3840 \times 2160$) through efficient token compression for DiT, significantly enhancing efficiency and enabling direct ultra-high-resolution image generation.
Sana~\cite{xie2024sana}, a pipeline for efficient and cost-effective training and synthesis of 4K images using a linear diffusion transformer, is capable of generating images at resolutions ranging from $1024 \times 1024$ to $4096 \times 4096$.
Sana~\cite{xie2024sana} introduces a deep compression VAE, \aka DC-AE~\cite{chen2024deep}, which compresses images with an aggressive down-sampling factor of $F=32$, thereby facilitating content creation at reduced cost.
Sana 1.5~\cite{xie2025sana}, building upon the original Sana, enables scaling from 1.6B to 4.8B parameters with significantly reduced computational resources, achieving scaling in both training and inference times for the linear diffusion transformer.

Despite significant improvements in resolution, both PixArt-$\Sigma$~\cite{chen2024pixart} and Sana~\cite{xie2024sana} primarily focus on the efficiency of image generation using token compression or linear attention mechanisms, leaving the potential of scalable MM-DiT models in 4K image synthesis unexplored.
Furthermore, these approaches overlook the high-frequency details and rich textures inherent in 4K images during both training and evaluation, which should be carefully considered, especially in the context of ultra-high-resolution image synthesis.
To bridge these gaps, we introduce the Diffusion-4K framework specifically designed to capture fine-grained visual details during training and incorporates novel evaluation metrics to quantify texture richness and detail fidelity.
\section{Aesthetic-4K Dataset}
\label{sec:aesthetic-4k}

To address the lack of a publicly available, high-quality 4K dataset, we introduce Aesthetic-4K, a meticulously curated dataset comprising standardized training and evaluation sets, namely Aesthetic-Train and Aesthetic-Eval, designed to support comprehensive research on ultra-high-resolution image synthesis, as detailed in \cref{sec:aesthetic-train} and \cref{sec:aesthetic-eval}.

\subsection{Aesthetic-Train}
\label{sec:aesthetic-train}

\begin{figure}
  \centering
  \includegraphics[width=\linewidth]{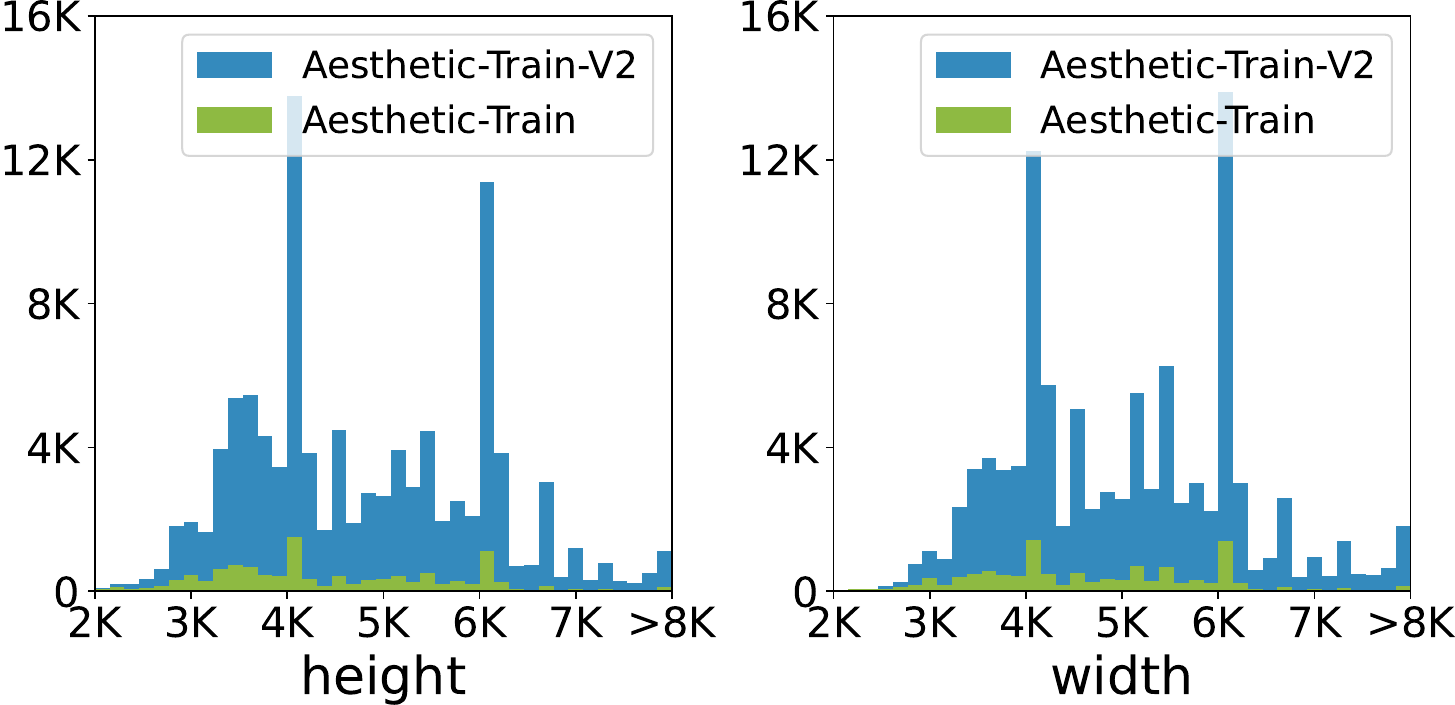}
  \caption{Histogram comparisons of image height and width in Aesthetic-Train~\cite{zhang2025diffusion4k} and Aesthetic-Train-V2. }
  \label{fig:aesthetic-train-comparison}
\end{figure}

\begin{table}
\centering
\caption{Statistical comparisons of Aesthetic-4K and PixArt-30K.}
\label{tab:dataset}
\resizebox{\columnwidth}{!}{
\begin{tabular}{c|c|c|c|c}
\toprule
Dataset & Median height & Median width & Average height & Average width \\
\midrule
PixArt-30K~\cite{chen2024pixart}  & 1615 & 1801 & 2531 & 2656 \\
\midrule
Aesthetic-Train~\cite{zhang2025diffusion4k}  & 4128 & 4640 & 4578 & 4838 \\
\midrule
Aesthetic-Train-V2  & 4605 & 5120 & 4861 & 5127 \\
\bottomrule
\end{tabular}
}
\end{table}

The Aesthetic-4K training set comprises high-quality images sourced from the Internet, carefully selected for their exceptional visual fidelity and fine details.
Simultaneously, precise and descriptive image captions are generated using the advanced GPT-4o model~\cite{hurst2024gpt}, ensuring strong alignment between visual content and language. 
Furthermore, we have rigorously filtered out low-quality images through manual inspection, excluding those with motion blur, focus issues, and mismatched text prompts, among other defects. 
The resulting curated images and corresponding captions constitute Aesthetic-Train, the training subset of Aesthetic-4K.

In addition to the previously proposed Aesthetic-Train~\cite{zhang2025diffusion4k}, which consists of 12,015 images, we have further established Aesthetic-Train-V2, comprising 105,288 high-quality image-text pairs.
The Aesthetic-Train-V2 is designed to evaluate the influence of scalable training data in ultra-high-resolution image synthesis, and is constructed using the same pipeline as Aesthetic-Train~\cite{zhang2025diffusion4k}, as previously described.
As illustrated in \cref{fig:aesthetic-train-comparison}, the introduced Aesthetic-Train-V2 demonstrates a substantial increase in training image volume for ultra-high-resolution image generation  compared to its predecessor. 
As detailed in \cref{tab:dataset}, the Aesthetic-Train has median image dimensions of 4128 pixels in height and 4640 pixels in width, while the Aesthetic-Train-V2 features even larger median dimensions of 4605 and 5120 pixels, respectively. 
Both training sets represent a substantial advancement over the open-source PixArt-30k~\cite{chen2024pixart}, which has notably smaller median dimensions of 1615 and 1801 pixels.

\begin{figure*}
\begin{center}
\subfloat[Image-text samples in training set.]{
\centering
\includegraphics[width=.484\textwidth]{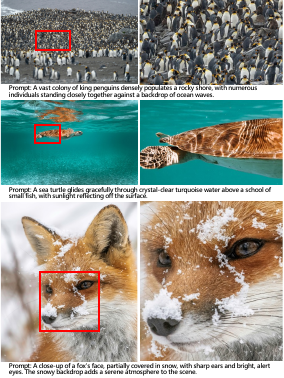}
\label{fig:demo_train}
}
\hfill
\subfloat[Image-text samples in evaluation set.]{
\centering
\includegraphics[width=.484\textwidth]{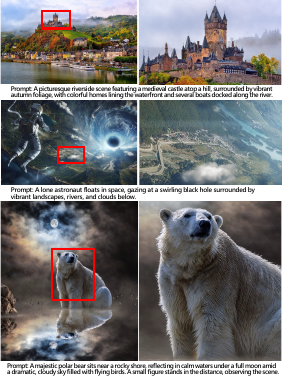}
\label{fig:demo_eval}
}
\caption{Illustration of image-text samples in the Aesthetic-4K dataset, which includes high-quality images and precise text prompts generated by GPT-4o, distinguished by exceptional visual quality and fine details.  } 
\label{fig:aesthetic-4k}
\end{center}
\end{figure*}

\subsection{Aesthetic-Eval}
\label{sec:aesthetic-eval}

For the evaluation set, termed Aesthetic-Eval, we select image-text pairs from the LAION-Aesthetics V2 6.5+ dataset, based on the criterion that the shorter side of each image exceeds 2048 pixels.
The LAION-Aesthetics dataset comprises 625,000 image-text pairs with predicted aesthetic scores of 6.5 or higher, as derived from LAION-5B~\cite{schuhmann2022laion}.
To mitigate the risk of overfitting in comprehensive assessments, we deliberately exclude any samples collected from the Internet when constructing the evaluation set.
The Aesthetic-Eval set comprises 2,781 high-quality images.
Among these, 195 images feature a short side exceeding 4096 pixels, forming a subset we denote as Aesthetic-Eval@4096.
Notably, only approximately $0.03\%$ of images in the LAION-Aesthetics V2 dataset meet the 4K resolution threshold, underscoring the scarcity of ultra-high-resolution samples in open-source datasets.
By introducing the Aesthetic-Eval, we establish a more appropriate benchmark for ultra-high-resolution image synthesis, advancing beyond the conventional $1024 \times 1024$ resolution typically used in prior evaluations~\cite{xie2024sana}.

In summary, the proposed Aesthetic-4K dataset covers a diverse range of categories that are highly relevant to real-world scenarios, including nature, travel, fashion, animals, film, art, food, sports, street photography, \etc. 
As illustrated in \cref{fig:aesthetic-4k}, we present several representative image-text pairs from both the training and evaluation sets of Aesthetic-4K, clearly demonstrating their exceptional quality.
\section{Methodology}

\begin{figure*}[ht]
\centering
\includegraphics[width=\linewidth]{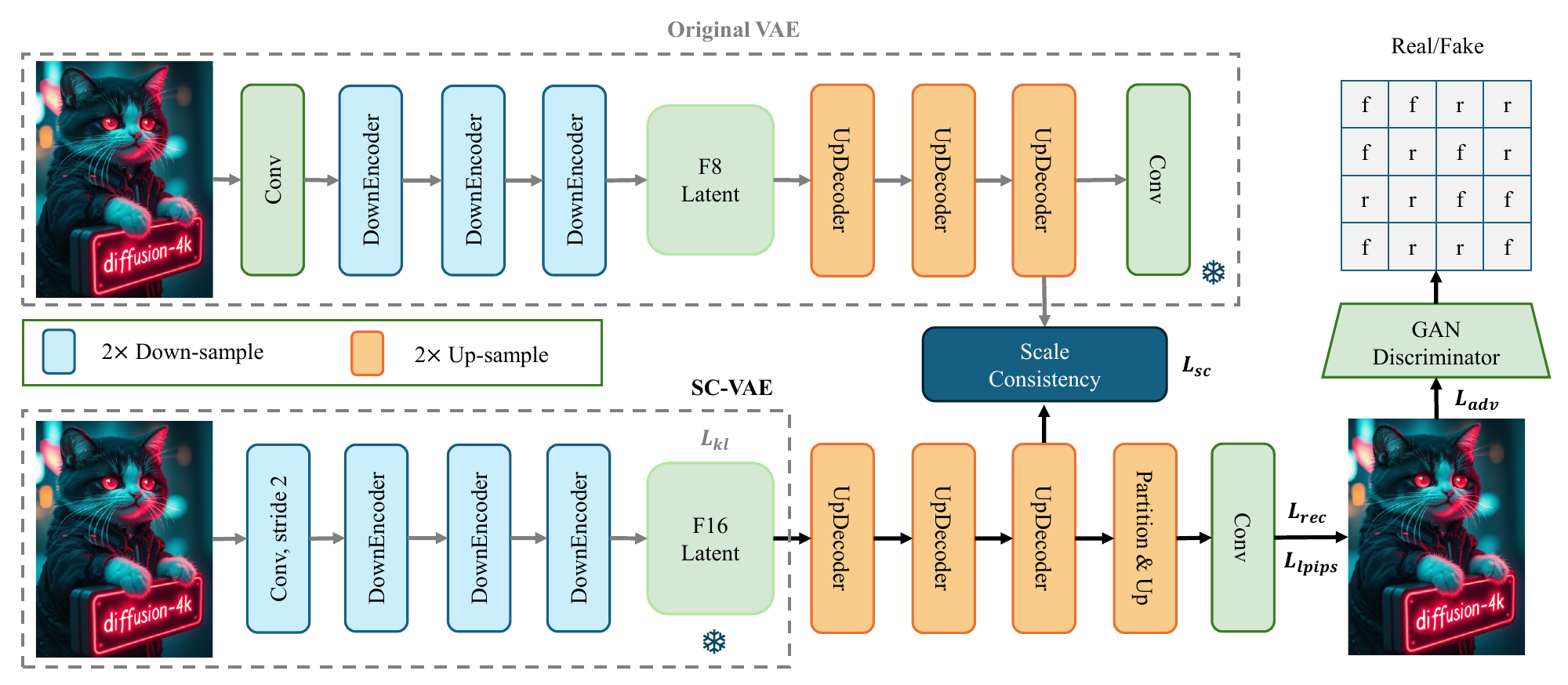}
\caption{The framework of the proposed SC-VAE. Our method shares the same latent space as the pre-trained latent diffusion model by fine-tuning only the decoder of the SC-VAE. }
\label{fig:framework_vae}
\end{figure*}

In this section, we propose Diffusion-4K, an efficient method specifically designed for various latent diffusion models, enabling direct training with photorealistic images at $4096 \times 4096$ resolution. 
The core advancements consist of two key components: SC-VAE and WLF, which are discussed in \cref{sec:sc_vae} and \cref{sec:wavelet}, respectively.

\subsection{Scale Consistent VAE}
\label{sec:sc_vae}

In latent diffusion models, the most commonly employed VAEs~\cite{esser2024scaling, Flux:2024:Online} with a down-sampling factor of $F=8$ encounter out-of-memory (OOM) issues during direct training and inference at extremely high resolution.
To mitigate this challenge, a partitioned VAE was proposed in our previous work~\cite{zhang2025diffusion4k}, offering a simple yet effective solution by increasing the down-sampling factor to $F=16$, thereby significantly reducing memory consumption.
Specifically, we apply a dilation rate of 2 in the first convolutional layer of the encoder $E$. 
In the final convolutional layer of the decoder $G$, we partition the input feature map, up-sample each partitioned segment by a factor of 2, apply the same convolution operator to each, and subsequently reorganize the outputs to form the final reconstruction.

In this section, we propose the Scale Consistent VAE (SC-VAE), which incorporates scale consistency regularization to enhance both reconstruction fidelity and generative performance, while maintaining the computational efficiency of the original partitioned VAE~\cite{zhang2025diffusion4k}. 
Formally, given a VAE consisting of an encoder $E$ and a decoder $G$, an input image $x$ is approximated by its reconstruction $\hat{x} = G(E(x))$.
As illustrated in \cref{fig:framework_vae}, the up-sampled feature map of the SC-VAE is calibrated with that of the original teacher VAE through self-distillation, formulating the \textbf{S}cale \textbf{C}onsistency (SC) loss as follows: 
\begin{equation}
  \mathcal{L}_{sc}(E, G) = \Vert  G_{o}^{L-1}(E_o(x)) - h(G_{sc}^{L-1}(E_{sc}(x))) \Vert ^2_2, 
  \label{eq:vae_scale_consistency}
\end{equation}
where $G_o^{L-1}(\cdot)$ and $G_{sc}^{L-1}(\cdot)$ represent the feature maps extracted from the penultimate decoder layer $L-1$ of the original VAE with $F=8$, and the SC-VAE with $F=16$, respectively, and $E_o$ and $E_{sc}$ are the encoders of the original VAE and the SC-VAE, respectively. The function $h(\cdot)$ denotes the up-sampling operation.
This regularization approach leverages the original VAE with $F=8$ as a teacher model to guide the optimization of the SC-VAE through consistency regularization in the feature maps, ensuring scale consistency between feature maps from VAEs with different down-sampling factors, thereby significantly enhancing the reconstruction and generation performance of the SC-VAE.
In addition to the scale consistency loss $\mathcal{L}_{sc}$ and the commonly used $L_2$ reconstruction loss $\mathcal{L}_{rec}$ and Kullback-Leibler (KL) loss $\mathcal{L}_{kl}$, we also incorporate perceptual loss $\mathcal{L}_{lpips}$~\cite{zhang2018unreasonable} and patch-based adversarial loss $\mathcal{L}_{adv}$~\cite{isola2017image}, which are widely adopted in fine-tuning VAE~\cite{esser2021taming, rombach2022high, chen2024deep}, to further improve the reconstruction quality. 
More precisely, a patch-based discriminator $D$ is introduced for adversarial training, which aims to differentiate between original and reconstructed images:
\begin{equation}
\begin{aligned}
  \mathcal{L}_{adv}(E, G, D) = [\log D(x) + \log (1-D(\hat{x}))].
  \label{eq:discriminator_loss}
\end{aligned}
\end{equation}
The adversarial approach facilitates to capture perceptually important local structures and improve local details.
Consequently, the total training objective for the SC-VAE is formulated as follows:
\begin{equation}
\begin{aligned}
  & \mathcal{L}_{vae} = \mathop{\min}\limits_{E, G} \mathop{\max}\limits_{D} \bigg[  \mathcal{L}_{rec}(E, G) + \lambda_{sc} \mathcal{L}_{sc}(E, G) + \lambda_{kl} \mathcal{L}_{kl}(E)  \\ & + \lambda_{lpips} \mathcal{L}_{lpips}(E, G) + \lambda_{adv} \frac{\nabla_{G^L}[\mathcal{L}_{lpips}]}{\nabla_{G^L}[\mathcal{L}_{adv}]} \mathcal{L}_{adv}(E, G, D)   \bigg], 
  \label{eq:vae_total_loss}
\end{aligned}
\end{equation}
where $\lambda_{sc}$, $\lambda_{kl}$, $\lambda_{lpips}$, and $\lambda_{adv}$ are the weights for the scale consistency loss $\mathcal{L}_{sc}$, KL loss $\mathcal{L}_{kl}$, perceptual loss $\mathcal{L}_{lpips}$, and patch-based adversarial loss $\mathcal{L}_{adv}$, respectively, and $\nabla_{G^L}[\cdot]$ denotes the gradient of its input \wrt the last layer $L$ of the decoder $G$.
The adaptive term $\frac{\nabla_{G^L}[\mathcal{L}_{lpips}]}{\nabla_{G^L}[\mathcal{L}_{adv}]}$ is calculated based on the gradients to balance the perceptual and adversarial loss.

Notably, in practice, our method maintains consistency in the latent space of the pre-trained latent diffusion model by fine-tuning only the decoder $G$ of the SC-VAE, resulting in the following optimization objective:
\begin{equation}
\begin{aligned}
  & \mathcal{L}_{vae}^{G} = \mathop{\min}\limits_{G} \mathop{\max}\limits_{D} \bigg[  \mathcal{L}_{rec}(G) + \lambda_{sc} \mathcal{L}_{sc}(G) \\ & + \lambda_{lpips} \mathcal{L}_{lpips}(G) + \lambda_{adv} \frac{\nabla_{G^L}[\mathcal{L}_{lpips}]}{\nabla_{G^L}[\mathcal{L}_{adv}]} \mathcal{L}_{adv}(G, D)  \bigg].
  \label{eq:vae_total_loss_wrt_G}
\end{aligned}
\end{equation}
This approach prevents distribution shifts in the latent space, thereby ensuring seamless compatibility with various diffusion models.

\subsection{Wavelet-based Latent Fine-tuning}
\label{sec:wavelet}

Wavelet transform has shown considerable success in image processing, primarily for decomposing low-frequency approximations and high-frequency details in images or features~\cite{guth2022wavelet, phung2023wavelet}.
In this section, we propose wavelet-based latent fine-tuning for diffusion models, which focuses on emphasizing high-frequency components while preserving low-frequency information, thereby significantly enhancing rich textures and fine details in 4K image generation.

Diffusion models~\cite{ho2020denoising, song2020denoising, song2020score, yang2023diffusion} consist of two Markov chains: a forward process that progressively perturbs data to noise, and a reverse process that recovers data from noise.
The forward process is typically hand-designed to gradually transform an arbitrary data distribution into a simple prior distribution (\eg, standard Gaussian), while the reverse process learns to invert this transformation by estimating the transition kernels using deep neural networks.
Formally, given a data distribution $\bm{x}_0 \sim q(\bm{x}_0)$ and standard Gaussian noise $\bm{\epsilon} \sim \mathcal{N}(\bm{0},\bm{I})$, the forward process gradually adds Gaussian noise to the data according to a discrete variance schedule $\{ \beta_t \in (0, 1) \}_{t=1}^T$:
\begin{equation}
  q(\bm{x}_t | \bm{x}_{t-1}) = \mathcal{N}(\bm{x}_t; \sqrt{1-\beta_t}\bm{x}_{t-1}, \beta_t\bm{I}).
  \label{eq:forward_process}
\end{equation}
By accumulating noise over time, we obtain:
\begin{equation}
  q(\bm{x}_t | \bm{x}_0) = \mathcal{N}(\bm{x}_t; \sqrt{\bar{\alpha}_t}\bm{x}_0, (1-\bar{\alpha}_t)\bm{I}),
  \label{eq:forward_chain}
\end{equation}
where $\alpha_t \coloneqq 1 - \beta_t$ and $\bar{\alpha}_t \coloneqq \prod_{s=1}^{t} \alpha_t$.
In the reverse process, the learnable transition kernel is modeled as:
\begin{equation}
  p_{\theta}(\bm{x}_{t-1} | \bm{x}_t) = \mathcal{N}(\bm{x}_{t-1}; \bm{\mu}_{\theta}(\bm{x}_t, t), \bm{\Sigma}_{\theta}(\bm{x}_t, t)),
  \label{eq:reverse_process}
\end{equation}
where the mean $\bm{\mu}_{\theta}(\bm{x}_t, t)$ and the variance $\bm{\Sigma}_{\theta}(\bm{x}_t, t)$ are parameterized by a denoising network $\theta$.
The standard training objective in diffusion models is to predict the added noise~\cite{ho2020denoising}, defined as:
\begin{equation}
  \mathcal{L}_{dm}(\theta) = \mathbb{E}_{t, \bm{x}_0, \bm{\epsilon}} \left[ \Vert \bm{\epsilon} - \bm{\epsilon}_\theta(\bm{x}_t, t)  \Vert^2 \right],
  \label{eq:noise_prediction}
\end{equation}
where $\bm{x}_t=\sqrt{\bar{\alpha}_t}\bm{x}_0 + \sqrt{1-\bar{\alpha}_t}\bm{\epsilon}$ and $t \sim \mathcal{U} \{1, \cdots,  T\}$.  
Here, $\mathcal{U}\{1, \cdots,  T\}$ denotes uniform sampling from the discrete timestep set $\{1, \cdots, T\}$.

Recent state-of-the-art approaches, such as SD3~\cite{esser2024scaling} and Flux~\cite{Flux:2024:Online}, adopt rectified flows~\cite{liu2022flow} to predict a velocity vector $\bm{v}$ that learns a straightforward transport mapping from the noise $\bm{\epsilon}$ to the data $\bm{x}_0$.
Given the linear interpolation $\bm{x}_t=(1-t)\bm{x}_0 + t\bm{\epsilon}$ where $t \sim \mathcal{U}(0,1)$, the training objective is formulated as follows:
\begin{equation}
  \mathcal{L}_{rf}(\theta) = \mathbb{E}_{t, \bm{x}_0, \bm{\epsilon}} \left[ w_t \Vert \bm{u}_t(\bm{x}_t) - \bm{v}_\theta(\bm{x}_t, t)  \Vert^2 \right],
  \label{eq:rectified_flow}
\end{equation}
where $\bm{u}_t(\bm{x}_t) = \frac{\mathrm{d} \bm{x}_t}{\mathrm{d} t} = \bm{\epsilon} - \bm{x}_0$, and $w_t$ denotes a time-dependent loss weighting factor.
To further enhance high-frequency details while preserving low-frequency approximations, we explicitly decompose latent features into the low- and high-frequency components using wavelet transform, resulting in the formulation of the \textbf{W}avelet-based \textbf{L}atent \textbf{F}ine-tuning (WLF) objective:
\begin{equation}
  \mathcal{L}_{wlf}(\theta) = \mathbb{E}_{t, \bm{x}_0, \bm{\epsilon}} \left[ w_t \Vert f(\bm{u}_t(\bm{x}_t)) - f(\bm{v}_\theta(\bm{x}_t, t)) \Vert^2 \right],
  \label{eq:wavelet_rectified_flow}
\end{equation}
where $f(\cdot)$ denotes discrete wavelet transform (DWT).
Notably, we utilize the Haar wavelet, widely adopted in real-world applications due to its efficiency.
Specifically, $L= \frac{1}{\sqrt{2}}\left[1, 1\right]$ and $H= \frac{1}{\sqrt{2}}\left[-1, 1\right]$ denote the low-pass and high-pass filters, which are used to construct four kernels in DWT with a stride of $2$, namely $LL^T, LH^T, HL^T, HH^T$.
The DWT kernels are then employed to decompose the input features into four sub-bands, the low-frequency approximation $\bm{x}_t^{ll}$ and high-frequency components $\bm{x}_t^{lh}, \bm{x}_t^{hl}, \bm{x}_t^{hh}$.

As illustrated in \cref{eq:wavelet_rectified_flow}, WLF decomposes the latent features into high- and low-frequency components, allowing the model to refine details (high-frequency) while maintaining the overall structure (low-frequency).
This decomposition not only enhances the capability to generate fine details but also ensures that the changes do not disrupt the underlying patterns, making the fine-tuning process both efficient and precise. 
Consequently, both low-frequency information and high-frequency details are incorporated into the WLF objective, contributing to a comprehensive optimization of 4K image synthesis. 

Moreover, our method supports various diffusion models by simply substituting the reconstruction objective, enabling seamless integration with conventional noise prediction approaches.
\section{Evaluation}
\label{sec:evaluation}

Existing automated evaluation metrics~\cite{heusel2017gans, hessel2021clipscore, schuhmann2022laion, xu2024imagereward} primarily focus on holistic evaluation and therefore fail to capture the highly structured textures and high-frequency details present in local patches of 4K imagery.
In this section, we introduce novel quantifiable indicators for assessing rich textures and fine details at the local patch level, demonstrating superior alignment with human perceptual preferences.
Furthermore, we present a comprehensive and multifaceted evaluation framework for ultra-high-resolution image generation that incorporates both holistic and local measures.

\subsection{Quantifiable Local Measures}

\emph{Emphasis on human-centric perceptual cognition:} 
The objective of our local indicators is to investigate the key factors influencing human perception of ultra-high-resolution images at the patch level and to establish quantifiable metrics that align closely with human evaluation.
Drawing on insights from perceptual psychology literature, we recognize that highly structured textures play a pivotal role in human visual cognition~\cite{julesz1981textons, stockwell2020texture, bergen1988early}. 
Accordingly, we propose innovative indicators to assess the richness of such textures and fine details at the patch level, including the GLCM Score and the image compression ratio with discrete cosine transform (DCT).
Given the sensitivity of human vision to local structural variations, the GLCM effectively captures diverse textural patterns, optical flow, and distortions through spatial interactions among neighboring pixels, thereby providing a representative characterization of human perceptual responses~\cite{gadkari2004image}.
This metric is well-aligned with human perceptual sensitivities, making it particularly suitable for evaluating texture richness in ultra-high-resolution imagery.
In parallel, the DCT-based image compression ratio offers a complementary perspective for assessing the preservation of intricate visual details in ultra-high-resolution images. 
Specifically, the GLCM Score is formulated as follows:
\begin{equation}
\label{eq:glcm_score}
s_{glcm} = - \frac{1}{P} \sum_{p=1}^{P} H(g_p),
\end{equation}
where $H$ represents entropy, and $g_{p}$ denotes the GLCM~\cite{haralick1973textural} derived from the local patch $p$ in the original image with 64 gray levels, defined by the radius $\delta=[1,2,3,4]$ and orientation $\theta=[0^{\circ}, 45^{\circ}, 90^{\circ}, 135^{\circ}]$.
In practice, we partition the gray image into $P$ local patches of size 64, and compute the average GLCM Score based on these partitioned local patches. 
Regrading the Compression Ratio, it is calculated as the ratio of the original size $M_o$ in memory to the compressed size $M_{c}$, \ie
\begin{equation}
\label{eq:compression_ratio}
s_{cr} = \frac{M_o}{M_{c}},
\end{equation}
where $M_{c}$ is obtained using the JPEG algorithm at a quality setting of 95.

\begin{figure}
\centering
\includegraphics[width=\linewidth]{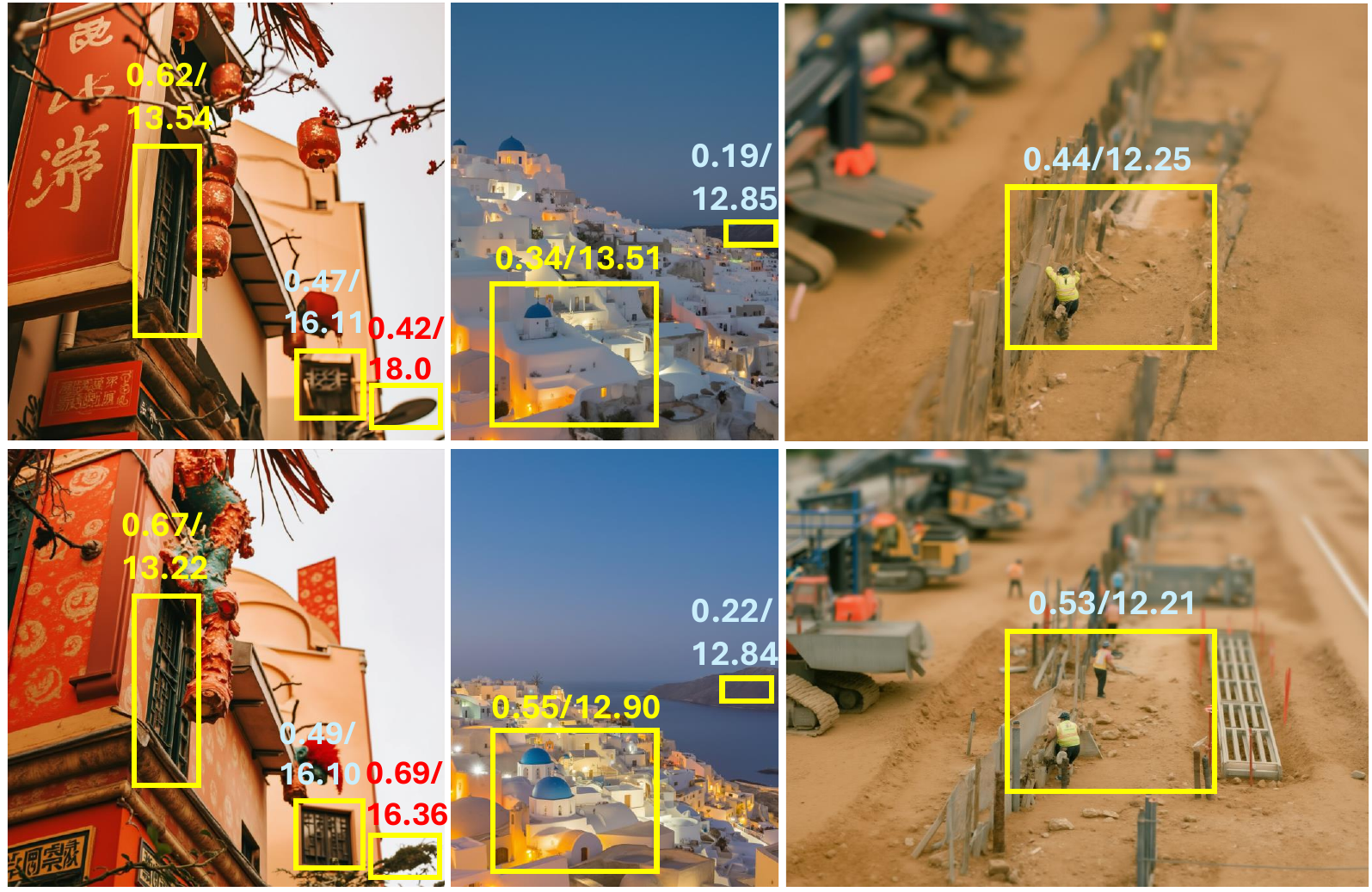}
\caption{Qualitative analysis of GLCM Score$\uparrow$ / Compression Ratio$\downarrow$. 
The top and bottom images are generated using the same prompts and random seed, but with different models.
Our indicators demonstrate a strong alignment with human-centric perceptual cognition of rich textures and fine details at the local patch level. }
\label{fig:image_align_with_glcm}
\end{figure}

\begin{table}
\begin{center}
\caption{Correlation with human evaluation. 
Our indicators exhibit superior alignment with human ratings compared to no-reference image quality assessment metrics MUSIQ~\cite{ke2021musiq} and MANIQA~\cite{yang2022maniqa}.
}
\label{tab:metrics}
\resizebox{\linewidth}{!}{
\begin{tabular}{c|c|c|c|c}
\toprule
Metric & GLCM Score & Compression Ratio & MUSIQ & MANIQA \\
\midrule
SRCC  & 0.75 & 0.53 & 0.36 & 0.20 \\
\midrule
PLCC  & 0.77 & 0.56 & 0.41 & 0.26 \\
\bottomrule
\end{tabular}
}
\end{center}
\end{table}

Furthermore, to demonstrate the alignment of our proposed local indicators with human perceptual preferences, we conduct a quantitative analysis on a diverse set of generated images and provide qualitative illustrations in \cref{fig:image_align_with_glcm}.
Additionally, as shown in \cref{tab:metrics}, we calculate the Spearman Rank-order Correlation Coefficient (SRCC) and the Pearson Linear Correlation Coefficient (PLCC) based on human evaluations of various patches sampled from the generated images.
These results indicate superior alignment with human ratings when compared to existing no-reference image quality assessment metrics, such as MUSIQ~\cite{ke2021musiq} and MANIQA~\cite{yang2022maniqa}.
In practice, five participants are asked to rate the extracted patches on a scale from 1 to 10 based on visual details, and the average scores are used to compute SRCC and PLCC, thereby mitigating inductive bias due to individual perceptual differences.
Our local indicators are thus specifically designed to ensure that performance metrics are meaningfully aligned with human perceptual judgments and cognitive processes.

\subsection{Multifaceted Assessment} 
Image quality assessment is a long-standing research topic. 
Recent approaches have introduced human preferences to evaluate the quality of generated images, training models to predict human ratings with single scalar score~\cite{xu2024imagereward, kirstain2023pick, wu2023human}. 
However, RAHF~\cite{liang2024rich} highlights the importance of fine-grained, multi-dimensional evaluations, emphasizing that such assessments offer greater interpretability and attribution, thus yielding a more comprehensive understanding of image quality compared to single-value metrics.


To facilitate a thorough evaluation for ultra-high-resolution image synthesis, we incorporate both conventional holistic metrics commonly used in deep generative models and our proposed local indicators.
Holistic measures, including FID~\cite{heusel2017gans}, Aesthetics~\cite{schuhmann2022laion}, and CLIPScore~\cite{hessel2021clipscore}, which have demonstrated effective in evaluating specific aspects
of generative model performance, are employed to provide an intuitive understanding of image synthesis in terms of generative quality, visual aesthetics and prompt adherence from a global perspective. 
Complementarily, quantitative local metrics are introduced to evaluate the rich textures and fine details at the patch level of 4K images.
These include the GLCM Score and the Compression Ratio, which align closely with human perceptual sensitivities and address an underexplored aspect of image quality assessment in ultra-high-resolution settings.
Together, these holistic and local measures form a comprehensive, multidimensional evaluation framework for ultra-high-resolution image synthesis.

\section{Experiments}

To demonstrate the effectiveness of our method, we conduct experiments with state-of-the-art latent diffusion models at various scales, including open-source SD3-2B~\cite{esser2024scaling}, and Flux-12B~\cite{Flux:2024:Online}. 
Specifically, we report mainstream evaluation metrics, such as FID~\cite{heusel2017gans}, Aesthetics~\cite{schuhmann2022laion} and CLIPScore~\cite{hessel2021clipscore}, along with the proposed GLCM Score and Compression Ratio metrics, for comprehensive assessments.
Additionally, we present both quantitative and qualitative results that highlight the ultra-high-resolution image reconstruction and generation capabilities of SC-VAE and WLF, respectively.
Finally, we conduct scalability analysis and comprehensive ablation studies to further validate the effectiveness of our approach.

\begin{table*}
  \centering
  \caption{Designed prompts for image caption and preference study with GPT-4o.}
  \label{tab:gpt_caption}
  \begin{tabular}{c|c}
    \toprule
    Tasks & Prompts \\ 
    \midrule
    \makecell[c]{Image \\ Caption} & \makecell[c]{ \{``\textbf{text}'': ``Directly describe with brevity and as brief as possible the scene or characters without any introductory phrase \\ like  `This image shows',  `In the scene',  `This image depicts' or similar phrases.  Just start describing the scene please.'' \} } \\
    \midrule
    \makecell[c]{Preference \\ Study} & \makecell[c]{ \{``\textbf{system}'': ``As an AI visual assistant, you are  analyzing two specific images.  When presented  with \\ a specific caption, it is required to  evaluate visual aesthetics,  prompt coherence  and fine details.'', \\ 
    ``\textbf{text}'': ``The caption for the two images is: $\langle$prompt$\rangle$.  Please answer the following questions: \\
    \emph{1. Visual Aesthetics}: Given the prompt, which image is of  higher-quality and aesthetically more pleasing? \\
    \emph{2. Prompt Adherence}: Which image looks more representative  to the text shown above and faithfully follows it? \\
    \emph{3. Fine Details}: Which image more accurately represents the fine visual details? Focus on clarity, \\  sharpness, and texture.  Assess the fidelity of fine elements such as edges, patterns,  and nuances in color.  \\ The more precise representation of these details is preferred! Ignore other aspects. \\
    Please respond me strictly in the following format: \\
    \emph{1. Visual Aesthetics}: $\langle$the first image is better$\rangle$ or  $\langle$the second image is better$\rangle$. The reason is $\langle$give your reason here$\rangle$. \\
    \emph{2. Prompt Adherence}: $\langle$the first image is better$\rangle$ or $\langle$the second image is better$\rangle$. The reason is $\langle$give your reason here$\rangle$. \\
    \emph{3. Fine Details}: $\langle$the first image is better$\rangle$ or  $\langle$the second image is better$\rangle$. The reason is $\langle$give your reason here$\rangle$. "\} }  \\
    \bottomrule
  \end{tabular}
\end{table*}

\subsection{Implementation Details}

We provide the training details for the two core components in our framework, including SC-VAE and WLF, respectively.

\noindent\textbf{Training Details of SC-VAE.} 
We fine-tune the SC-VAE on the SA-1B dataset~\cite{kirillov2023segment} for one epoch with a batch size of $256$ and employ EMA weights. 
For pre-processing, the images are resized and randomly cropped to $512 \times 512$ resolution. 
The SC-VAE and GAN discriminator are trained with a constant learning rate of $1 \times 10^{-5}$ and weight decay of $1 \times 10^{-4}$.
The loss weights $\lambda_{lpips}$, $\lambda_{adv}$ and $\lambda_{sc}$ in \cref{eq:vae_total_loss_wrt_G} are set to 0.1, 0.05, and 1.0, respectively.
Note that only the decoder of the SC-VAE is fine-tuned during the training phase to maintain consistency in the latent space.

\noindent\textbf{Training Details of WLF.} 
During pre-processing, images are resized to a shorter dimension of 4096, randomly cropped to a $4096 \times 4096$ resolution, and normalized with a mean and standard deviation of 0.5. 
The SC-VAE compresses the pixel space $\mathbb{R}^{H \times W \times 3}$ into a latent space $\mathbb{R}^{\frac{H}{F} \times \frac{W}{F} \times C}$, where $F=16$. 
The encoded latents are normalized using the mean and standard deviation from the pretrained latent diffusion models, which are globally computed over a subset of the training data.
The latent diffusion models are then optimized using the WLF objective in \cref{eq:wavelet_rectified_flow}. 
Regarding the text encoder, both CLIP~\cite{radford2021learning} and T5-XXL~\cite{raffel2020exploring} serve as the default models for text comprehension in SD3~\cite{esser2024scaling} and Flux~\cite{Flux:2024:Online}. 
To conserve memory, text embeddings for latent diffusion models are pre-computed, thus eliminating the need to load text encoders into the GPU during the training phase. 
We employ a default patch size of $P=2$ for DiTs, including SD3-2B and Flux-12B.
The latent diffusion models are optimized using the WLF objective with all parameters unfrozen, whereas text encoders and the SC-VAE remain fixed during training. 
In practice, we use the AdamW~\cite{loshchilov2017decoupled} optimizer with a constant learning rate of $1 \times 10^{-6}$ and weight decay of $1 \times 10^{-4}$. 
We employ mixed-precision training with a batch size of 32 and use ZeRO Stage 2 with CPU offload techniques~\cite{rajbhandari2020zero, ren2021zero}. 
The fine-tuning of SD3-2B and Flux-12B is conducted on 2 A800-80G GPUs and 8 A100-80G GPUs, respectively, using the Aesthetic-Train-V2 dataset for 50K training steps.
Note that we use the open-source Flux.1-dev version trained with guidance distillation, and adopt the default guidance scale of 3.5 for WLF.

\noindent\textbf{Evaluation Details.} 
During evaluation on the established Aesthetic-Eval@2048 set, images are generated using a guidance scale of 7.0 by discretizing the ordinary differential equation (ODE) process with an Euler solver, employing 28 sampling steps for SD3-2B and 50 sampling steps for Flux-12B, respectively.
The FID~\cite{heusel2017gans} measures the similarity between two sets of images, typically between real and generated images, by comparing their feature distributions extracted by Inception v3 at a resolution of $299 \times 299$.
The CLIPScore~\cite{hessel2021clipscore} evaluates the semantic similarity between images and text descriptions using CLIP embeddings. 
The Aesthetics~\cite{schuhmann2022laion} score is predicted using a simple linear model on top of CLIP ViT-L/14.
The GLCM Score is calculated based on the partitioned local patches of size 64, and the Compression Ratio is determined using the JPEG algorithm at a quality setting of 95.

\noindent\textbf{Detailed Prompts for GPT-4o.} 
As depicted in \cref{tab:gpt_caption}, we provide the detailed prompts used for generating image captions with GPT-4o in the Aesthetic-4K dataset.
Additionally, we present the complete prompts used in the preference study with GPT-4o to evaluate AI preferences for generated images across different aspects, including visual aesthetics, prompt adherence, and fine details.

\subsection{Experimental Results}
\label{sec:experimental_results}

\begin{table}
  \centering
  \caption{Quantitative reconstruction results of SC-VAE with a down-sampling factor of $F=16$ on Aesthetic-Train at $4096 \times 4096$ resolution. 
  }
  \label{tab:vae_reconstruction}
  \resizebox{\linewidth}{!}{
  \begin{tabular}{c|c|c|c|c|c}
    \toprule
    Model & rFID $\downarrow$ & NMSE $\downarrow$ & PSNR $\uparrow$ & SSIM $\uparrow$ & LPIPS $\downarrow$ \\
    \midrule
    SD3-VAE-F16~\cite{zhang2025diffusion4k} & 1.40 & 0.09 & 28.82 & 0.76 & 0.15 \\ 
    SD3-VAE-F16-SC & \textbf{0.59} & \textbf{0.07} & \textbf{30.90} & \textbf{0.80} & \textbf{0.10} \\
    \midrule
    Flux-VAE-F16~\cite{zhang2025diffusion4k} & 1.69 & 0.08 & 29.22 & 0.79 & 0.16 \\ 
    Flux-VAE-F16-SC & \textbf{0.45} & \textbf{0.05} & \textbf{33.41} & \textbf{0.86} & \textbf{0.09} \\ 
    \bottomrule
  \end{tabular}
  }
\end{table}

\begin{figure}
\centering 
\subfloat[Original images and local patches.]{
\centering
\includegraphics[width=\linewidth]{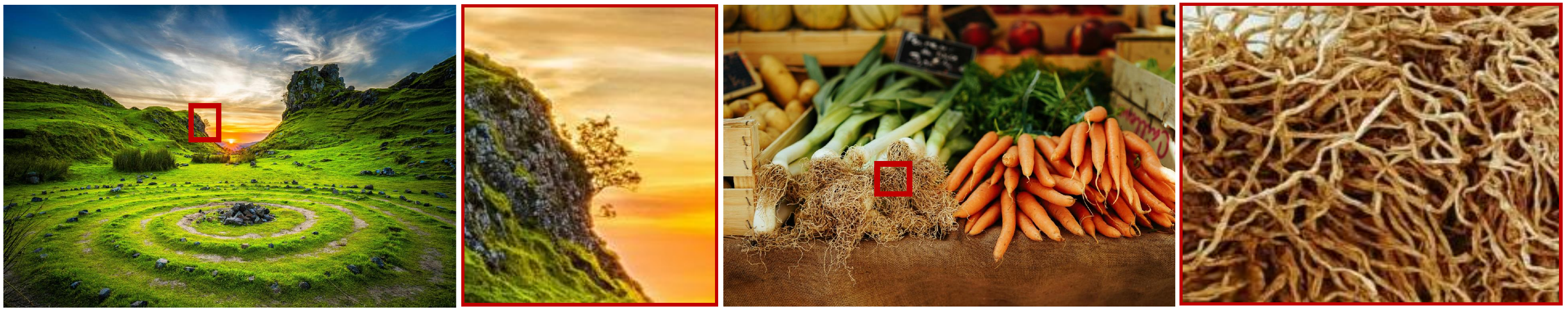}
\label{fig:vae_reconstruction_original}
}
\hfill
\subfloat[Reconstruction results by partitioned VAE~\cite{zhang2025diffusion4k}.]{
\centering
\includegraphics[width=\linewidth]{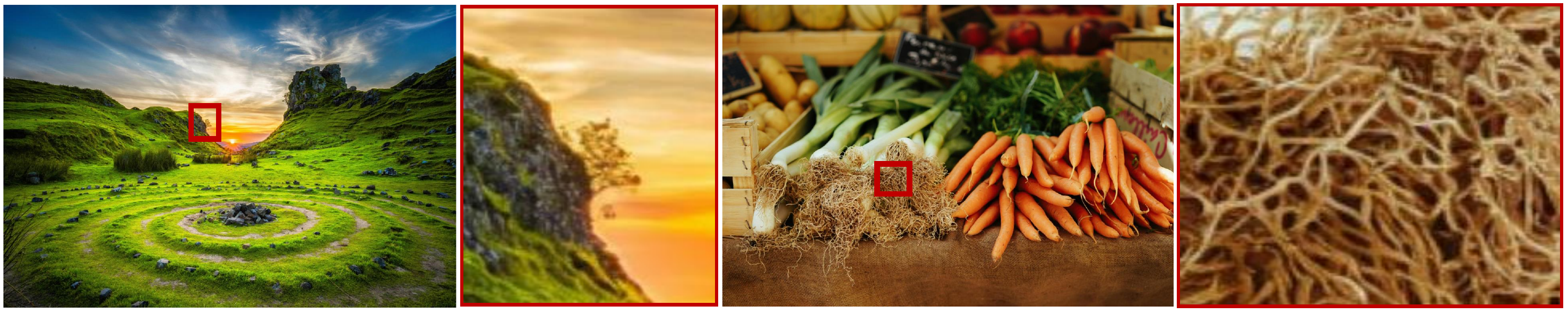}
\label{fig:vae_reconstruction_f16}
}
\hfill
\subfloat[Reconstruction results by SC-VAE.]{
\centering
\includegraphics[width=\linewidth]{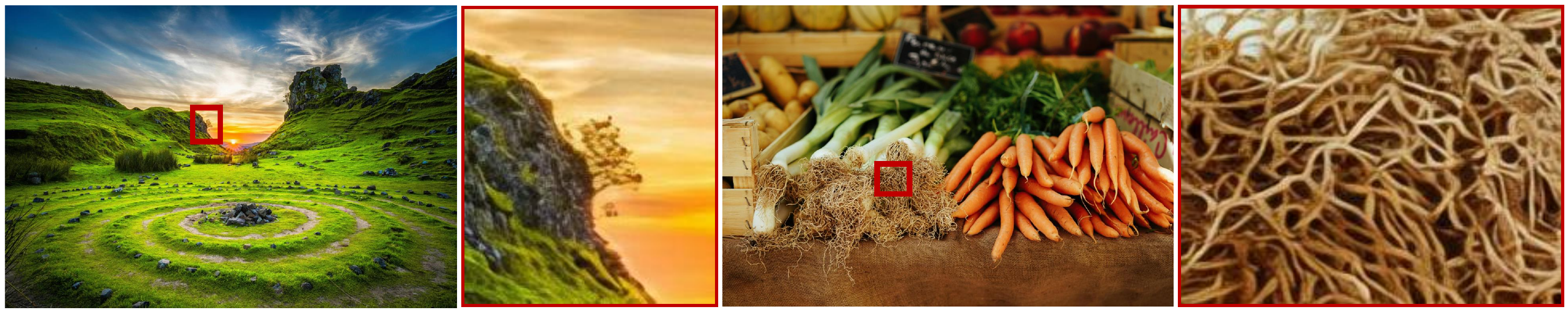}
\label{fig:vae_reconstruction_f16_sc}
}
\caption{Qualitative reconstruction comparisons of ultra-high-resolution images using partitioned VAE~\cite{zhang2025diffusion4k} and SC-VAE with a down-sampling factor of $F=16$.
}
\label{fig:vae_reconstruction}
\end{figure}

\noindent\textbf{Analysis of SC-VAE.}
As illustrated in \cref{tab:vae_reconstruction}, we report comprehensive evaluation results, including rFID, Normalized Mean Square Error (NMSE), Peak Signal-to-Noise Ratio (PSNR), Structural Similarity Index Measure (SSIM)~\cite{wang2004image}, and Learned Perceptual Image Patch Similarity (LPIPS)~\cite{zhang2018unreasonable}, to assess the reconstruction performance of SC-VAE on Aesthetic-Train at $4096 \times 4096$ resolution. 
We present detailed results of SC-VAEs in SD3 and Flux, using a down-sampling factor of $F=16$, along with baseline results in~\cite{zhang2025diffusion4k} without fine-tuning the decoder for comparison.
Additionally, we include visualizations of original images and local patches, reconstruction results by the partitioned VAE~\cite{zhang2025diffusion4k}, and results from the SC-VAE, as shown in \cref{fig:vae_reconstruction_original}, \cref{fig:vae_reconstruction_f16}, and \cref{fig:vae_reconstruction_f16_sc}, respectively.
The reconstruction results in \cref{fig:vae_reconstruction_f16_sc}, which incorporate scale consistency, exhibit enhanced detail in local patches compared to those in \cref{fig:vae_reconstruction_f16}.

\begin{figure}
\centering 
\subfloat[Generation results by partitioned VAE~\cite{zhang2025diffusion4k}.]{
\centering
\includegraphics[width=\linewidth]{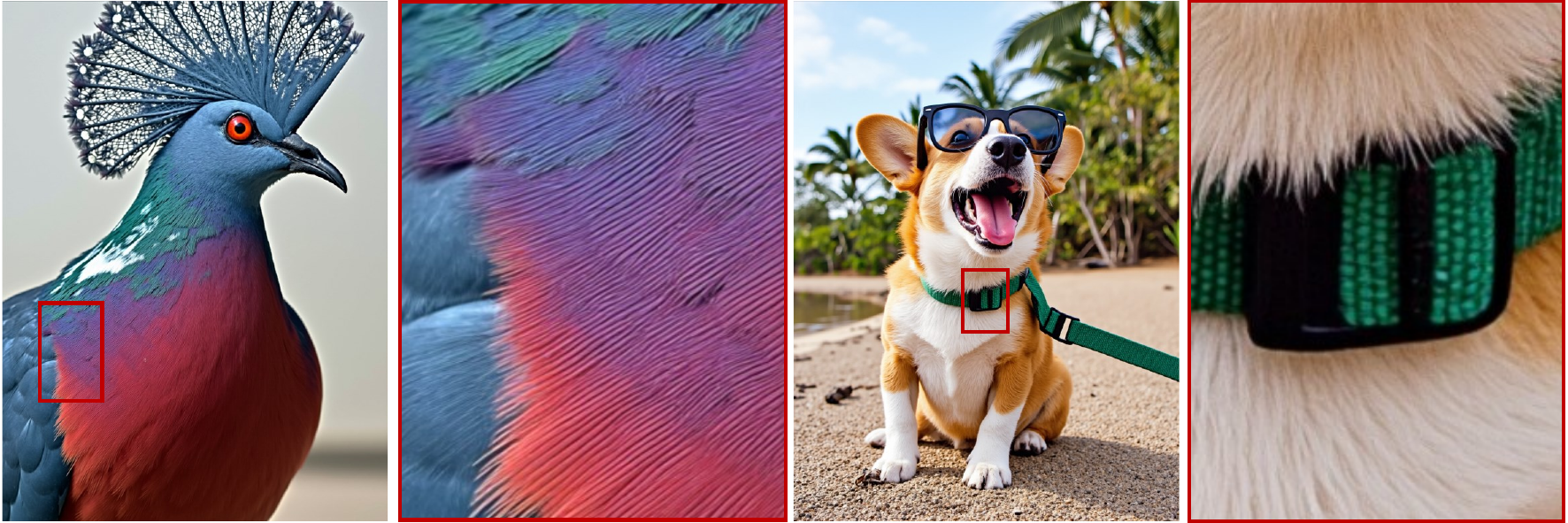}
\label{fig:vae_generation_f16}
}
\hfill
\subfloat[Generation results by SC-VAE.]{
\centering
\includegraphics[width=\linewidth]{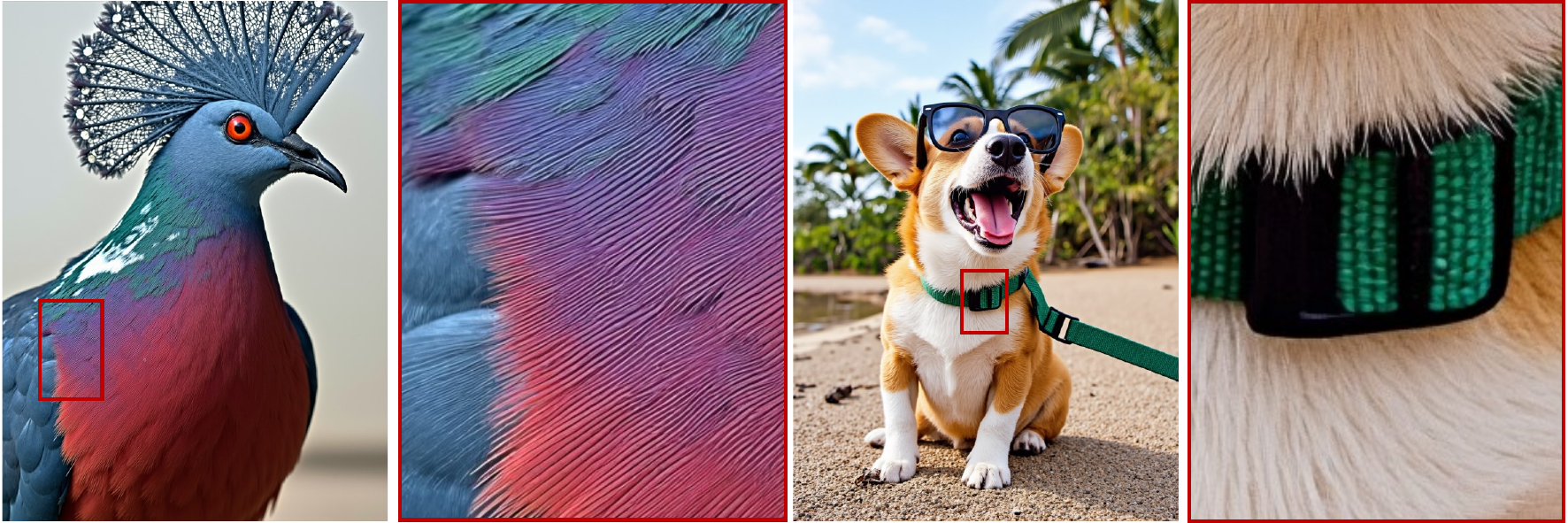}
\label{fig:vae_generation_f16_sc}
}
\caption{Qualitative generation comparisons of ultra-high-resolution images using partitioned VAE~\cite{zhang2025diffusion4k} and SC-VAE with a down-sampling factor of $F=16$.
}
\label{fig:vae_generation}
\end{figure}

Furthermore, in addition to reconstruction performance, we present qualitative ultra-high-resolution image generation results for comparison, including images synthesized with the latent diffusion model using the partitioned VAE~\cite{zhang2025diffusion4k} in \cref{fig:vae_generation_f16}, and those generated by the latent diffusion model using the SC-VAE in \cref{fig:vae_generation_f16_sc}. 
Note that the same random seeds and text prompts are employed to ensure a fair comparison.
Similarly, the generation results in \cref{fig:vae_generation_f16_sc}, which integrate scale consistency, show improved detail in local patches compared to those presented in \cref{fig:vae_generation_f16}.

Consequently, both quantitative and qualitative results demonstrate the effectiveness of our SC-VAE in ultra-high-resolution image reconstruction and generation.
Notably, our SC-VAE resolves the OOM issue encountered by the original VAE in 4K image generation, and the proposed scale consistency regularization approach significantly improves the reconstruction and generation performance of the partitioned VAE with $F=16$, while simultaneously preventing potential distribution shifts in the latent space.

\begin{table*}
  \centering
  \caption{Quantitative results of latent diffusion models on Aesthetic-Eval@2048 at $2048 \times 2048$ resolution. } 
  \label{tab:eval_2048_total}
  \begin{tabular}{c|c|ccc|cc}
    \toprule
    \multirow{2}{*}{Model} & \multirow{2}{*}{Architecture} & \multicolumn{3}{c|}{Holistic Measures} & \multicolumn{2}{c}{Local Measures} \\
    \cmidrule(lr){3-5} \cmidrule(lr){6-7}
    & & FID $\downarrow$ & CLIPScore $\uparrow$  & Aesthetics $\uparrow$ & GLCM Score $\uparrow$ & Compression Ratio $\downarrow$ \\ 
    \midrule 
    SD3-F16@2048~\cite{zhang2025diffusion4k} & \multirow{4}{*}{MM-DiT \& Sinusoidal PE} & 43.82 & 31.50 & 5.91 & 0.75 & 11.23 \\ 
    SD3-F16-WLF@2048~\cite{zhang2025diffusion4k} & & 40.18 & 34.04 & 5.96 & 0.79 & 10.51 \\ 
    SD3-F16-SC@2048 & & 38.93 & 33.98 & 6.06 & 0.79 & 10.34 \\
    SD3-F16-SC-WLF@2048 & & \textbf{37.83} & \textbf{34.98} & \textbf{6.14} & \textbf{0.80} & \textbf{10.28} \\ 
    \midrule
    Flux-F16@2048~\cite{zhang2025diffusion4k} & \multirow{4}{*}{MM-DiT \& RoPE} & 50.57 & 30.41 & 6.36 & 0.58 & 14.80 \\  
    Flux-F16-WLF@2048~\cite{zhang2025diffusion4k} & & 39.49 & 34.41 & 6.37 & 0.61 & 13.60 \\ 
    Flux-F16-SC@2048 & & 43.28 & 34.35 & 6.36 & 0.74 & 10.89 \\ 
    Flux-F16-SC-WLF@2048 & & \textbf{38.38} & \textbf{34.42} & \textbf{6.37} & \textbf{0.79} & \textbf{9.95} \\ 
    \midrule 
    PixArt-$\Sigma$@2048~\cite{chen2024pixart} & DiT \& Sinusoidal PE & 38.77 & 35.18 & 6.66 & 0.71 & 10.76 \\
    \midrule 
    Sana@2048~\cite{xie2024sana} & Linear-DiT \& Sinusoidal PE & 39.01 & 35.90 & 6.55 & 0.75 & 10.58 \\ 
    \bottomrule
  \end{tabular}
\end{table*}

\noindent\textbf{Quantitative Image Quality Assessment.}
Regarding image quality assessment, we perform comprehensive comparisons using mainstream evaluation metrics, such as FID~\cite{heusel2017gans}, Aesthetics~\cite{schuhmann2022laion} and CLIPScore~\cite{hessel2021clipscore}, to provide an intuitive understanding of holistic image quality and text prompt adherence.
As aforementioned, these holistic evaluation metrics are insufficient for comprehensive assessment of ultra-high-resolution image synthesis, particularly in evaluating the fine details of 4K images. 
To address this gap, we introduce additional comparisons using the GLCM Score, which assesses the texture richness of ultra-high-resolution images.
Simultaneously, we report the Compression Ratio using the JPEG algorithm at a quality setting of 95, which can serve as an important indicator to evaluate the preservation of fine details in image quality assessment.

As illustrated in \cref{tab:eval_2048_total}, we present experimental results on Aesthetic-Eval@2048 using various latent diffusion models, including SD3-2B and Flux-12B with the MM-DiT architecture.
These results demonstrate the effectiveness of our method, which incorporates both SC-VAE and WLF, in enhancing various aspects compared to the previous approach~\cite{zhang2025diffusion4k}, including generative image quality, prompt adherence and fine details, \etc.
Additionally, we provide quantitative comparisons with other direct ultra-high-resolution image synthesis approaches, including state-of-the-art diffusion models, such as PixArt-$\Sigma$~\cite{chen2024pixart} with Key-Value (KV) token compression, and Sana~\cite{xie2024sana} with linear attention transformer, which have already been trained on their private high-quality ultra-high-resolution datasets.
The quantitative results indicate that while PixArt-$\Sigma$~\cite{chen2024pixart} and Sana~\cite{xie2024sana} achieve superior visual aesthetics and prompt alignment, our method delivers higher generative quality, richer textures, and finer visual details.

\begin{table}
  \centering
  \caption{Memory consumption and inference speed of direct image synthesis at $4096 \times 4096$ resolution. The result is tested on one A100 GPU with BF16 Precision. }
  \label{tab:memory_and_speed}
  \begin{tabular}{l|c|c}
    \toprule
    Model & Memory & Time (s/step) \\
    \midrule
    SD3-F8@4096 & OOM  & -  \\ 
    SD3-F16-SC-WLF@4096  & 31.3GB  & 1.16  \\
    SD3-F16-SC-WLF@4096 (CPU offload) & 16.1GB  & 1.22  \\
    \midrule
    Flux-F8@4096 & OOM & -  \\ 
    Flux-F16-SC-WLF@4096  & 50.4 GB & 2.42  \\
    Flux-F16-SC-WLF@4096 (CPU offload)  & 26.9 GB & 3.16  \\ 
    \bottomrule
  \end{tabular}
\end{table}

\begin{figure*}[ht!]
  \centering
  \includegraphics[width=\linewidth]{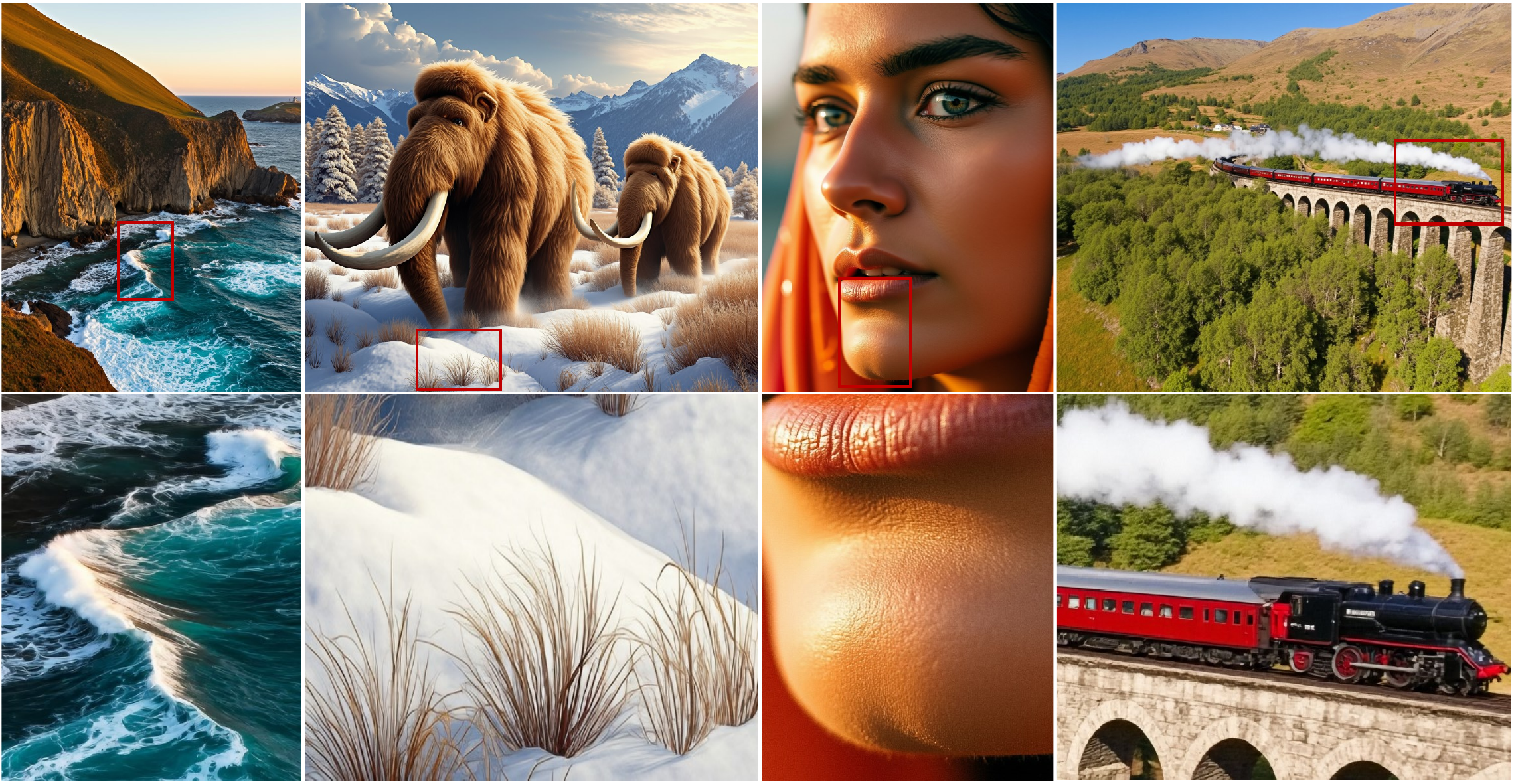}
  \caption{Qualitative results synthesized by our Diffusion-4K, emphasizing exceptional fine details in the generated 4K images.}
  \label{fig:demo}
\end{figure*}

\noindent\textbf{Qualitative Image Synthesis.}
As illustrated in \cref{fig:demo}, we present qualitative ultra-high-resolution images synthesized with Diffusion-4K using prompts from Sora~\cite{Sora:2024:Online}, powered by the state-of-the-art latent diffusion model, Flux-12B.
Although WLF fine-tunes the diffusion model at $4096 \times 4096$ resolution, our method is capable of synthesizing ultra-high-resolution images at various aspect ratios and resolutions.
The qualitative results prominently demonstrate the impressive performance of our approach in 4K image generation, with a particular emphasis on fine details.
Additionally, we report the inference details in \cref{tab:memory_and_speed}, which outline the time and memory consumption associated with our method for directly generating 4K images.

\begin{figure*}
\begin{center}
\subfloat[Qualitative results by our Diffusion-4K.]{
\centering
\includegraphics[width=\linewidth]{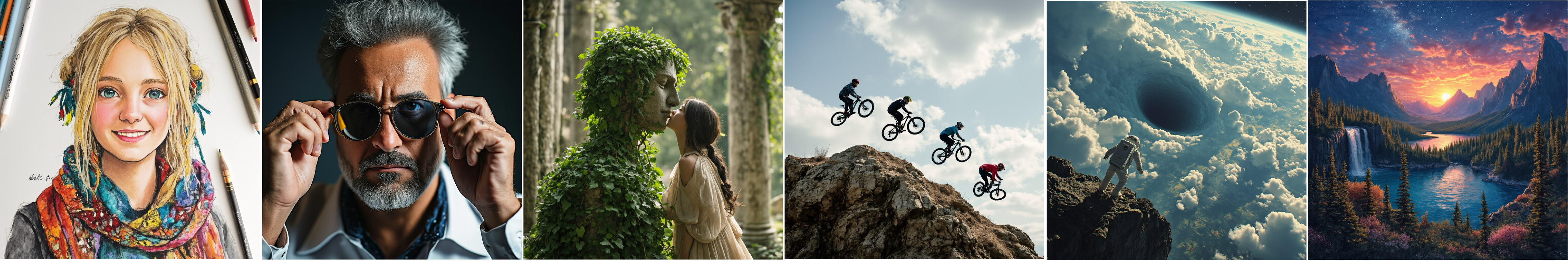}
\label{fig:qualitative_comparison_diffusion4k}
}
\hfill
\subfloat[Qualitative results by PixArt-$\Sigma$~\cite{chen2024pixart}.]{
\centering
\includegraphics[width=\linewidth]{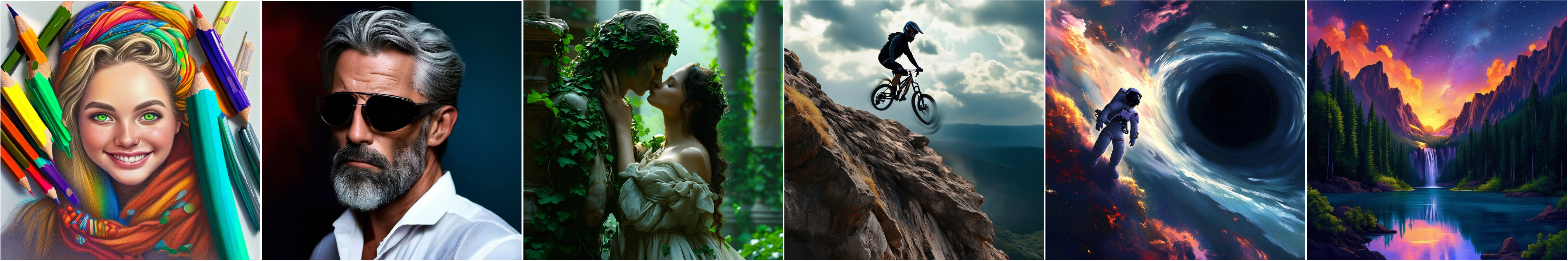}
\label{fig:qualitative_comparison_pixart}
}
\hfill
\subfloat[Qualitative results by Sana~\cite{xie2024sana}.]{
\centering
\includegraphics[width=\linewidth]{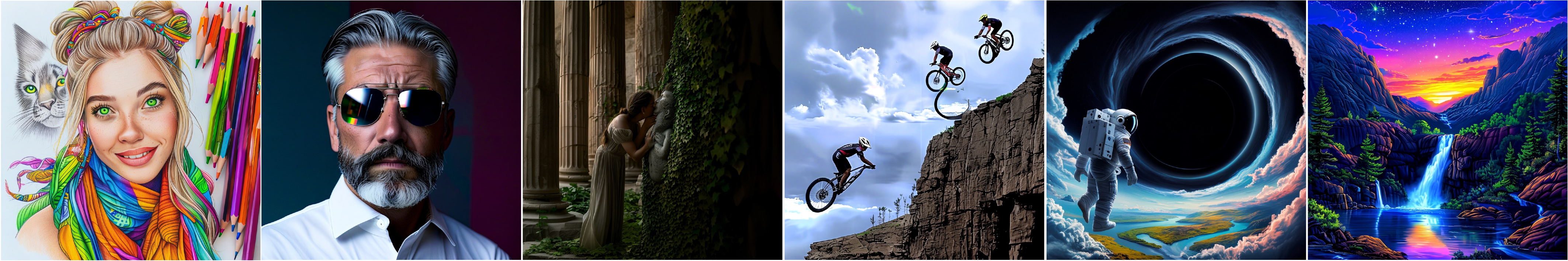}
\label{fig:qualitative_comparison_sana}
}
\caption{Qualitative results on Aesthetic-Eval@2048 at 2048 × 2048 resolution, including our proposed Diffusion-4K, PixArt-$\Sigma$~\cite{chen2024pixart} and Sana~\cite{xie2024sana}, respectively.  } 
\label{fig:qualitative_comparisons}
\end{center}
\end{figure*}

\begin{figure*}
  \centering
  \includegraphics[width=\linewidth]{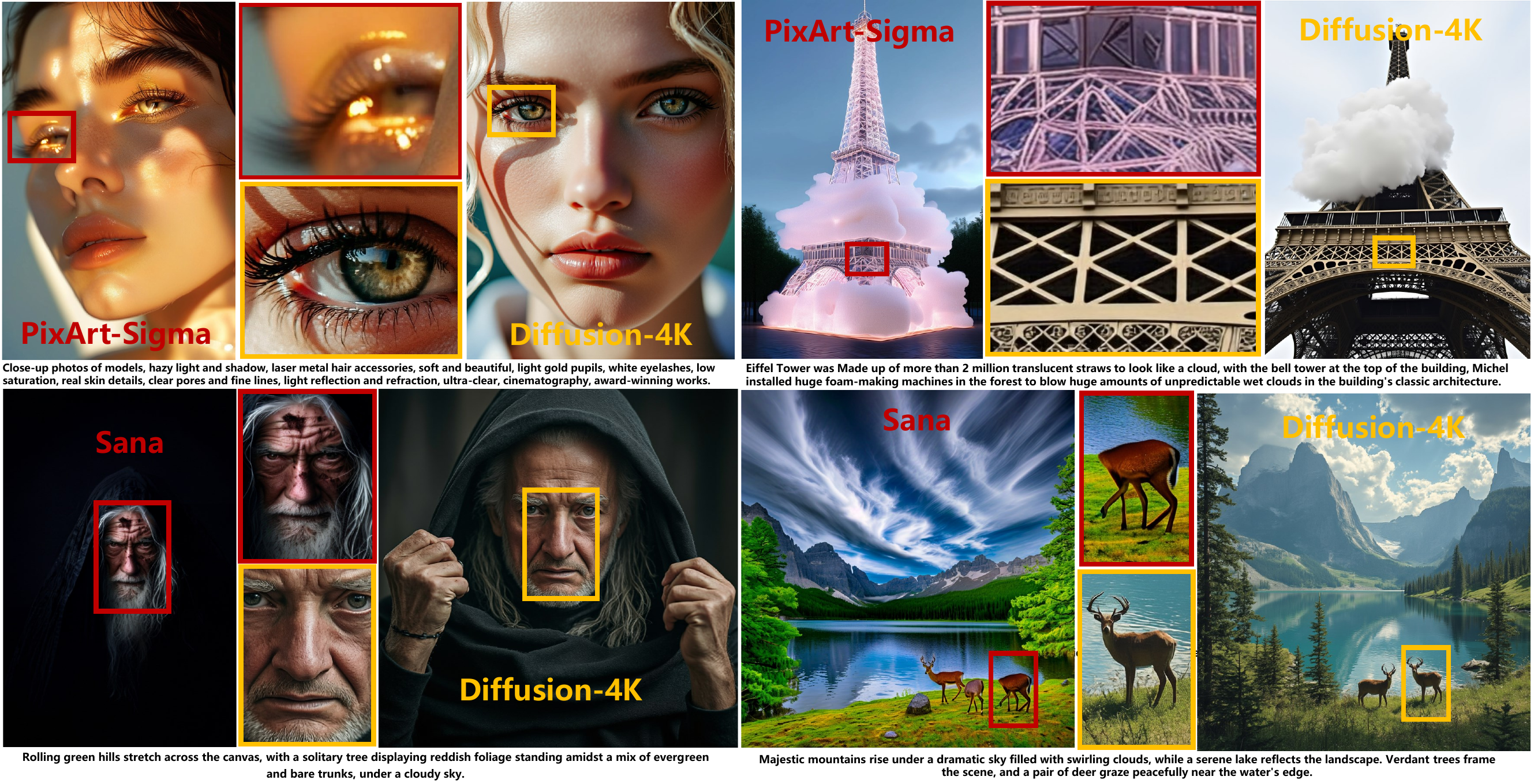}
  \caption{
  We present qualitative comparisons with PixArt-$\Sigma$\cite{chen2024pixart} and Sana\cite{xie2024sana} in local patches using identical prompts, where the images generated by PixArt-$\Sigma$ and Sana are shown on the left, and those synthesized by our Diffusion-4K are shown on the right.
  As illustrated by the yellow-highlighted patches compared to the red-highlighted ones, our method demonstrates clear superiority in rendering rich textures and intricate fine details.
  }
  \label{fig:demo_comparisons}
\end{figure*}

As illustrated in \cref{fig:qualitative_comparisons}, we present qualitative results on Aesthetic-Eval at $2048 \times 2048$ resolution using direct ultra-high-resolution image synthesis approaches, including our proposed Diffusion-4K, PixArt-$\Sigma$~\cite{chen2024pixart} and Sana~\cite{xie2024sana}, respectively.
To further highlight the strengths of our method in producing highly realistic images with rich textures and fine details, we also provide side-by-side qualitative comparisons of local image patches in \cref{fig:demo_comparisons}.
These comparisons clearly demonstrate that Diffusion-4K consistently outperforms PixArt-$\Sigma$ and Sana in rendering rich textures and fine details, as evidenced by the yellow-marked patches in contrast to the red-marked ones.

\begin{figure}
  \centering
  \includegraphics[width=\linewidth]{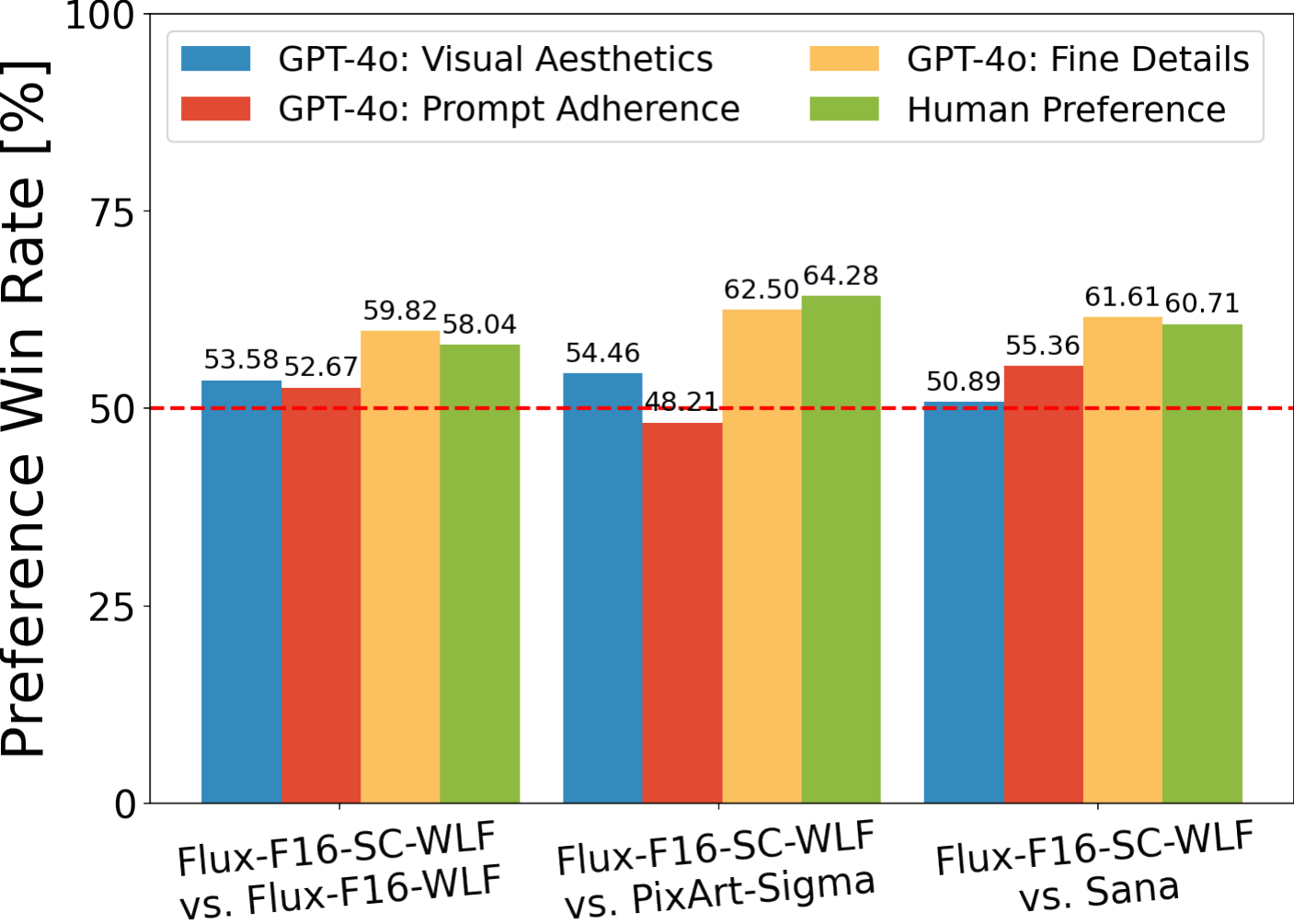}
  \caption{Human and GPT-4o preference evaluation. Our Flux-F16-SC-WLF model consistently outperforms Flux-F16-WLF~\cite{zhang2025diffusion4k} in terms of visual aesthetics, prompt adherence, fine details, and human preference, demonstrating the effectiveness of our approach.
  Furthermore, our method exhibits better human preferences compared to state-of-the-art models, including PixArt-$\Sigma$~\cite{chen2024pixart} and Sana~\cite{xie2024sana}.
  }
  \label{fig:win_rate}
\end{figure}

\noindent\textbf{Preference Study.}
To demonstrate the effectiveness of our method in ultra-high-resolution image synthesis, we perform both human and AI preference studies.
In the human preference study, participants rate pairwise outputs from two different latent diffusion models for comparison, including Flux-F16-SC-WLF \vs Flux-WLF-F16~\cite{zhang2025diffusion4k}, Flux-F16-SC-WLF \vs PixArt-$\Sigma$~\cite{chen2024pixart}, and Flux-F16-SC-WLF \vs Sana~\cite{xie2024sana}.
Ten participants rate their preferences for the generated images, with the average scores being used to mitigate inductive bias from individual differences in human ratings.
Additionally, for the AI preference study, we utilize the advanced multi-modal model, GPT-4o~\cite{hurst2024gpt}, as the evaluator. 
Detailed prompts, as outlined in \cref{tab:gpt_caption}, are employed in this evaluation. 
We conduct experiments with 112 text prompts sampled from Sora~\cite{Sora:2024:Online}, PixArt~\cite{chen2024pixart}, SD3~\cite{esser2024scaling}, \etc. 
As illustrated in \cref{fig:win_rate}, our method consistently achieves a higher win rate in both human and AI evaluations compared to its predecessor~\cite{zhang2025diffusion4k}, demonstrating improvements in visual aesthetics, prompt adherence, fine details, and overall human preference in ultra-high-resolution image generation.
Moreover, our method attains higher human preference scores than state-of-the-art models, including PixArt-$\Sigma$~\cite{chen2024pixart} and Sana~\cite{xie2024sana}.

\subsection{Scalability Analysis}
\label{sec:scalability_analysis}

\begin{table*}
  \centering
  \caption{Quantitative scalability results on Aesthetic-Eval@2048 at $2048 \times 2048$ resolution. } 
  \label{tab:ablation_scaling_data}
  \begin{tabular}{c|c|ccc|cc}
    \toprule
    \multirow{2}{*}{Model} & \multirow{2}{*}{Training set} & \multicolumn{3}{c|}{Holistic Measures} & \multicolumn{2}{c}{Local Measures} \\
    \cmidrule(lr){3-5} \cmidrule(lr){6-7}
    & & FID $\downarrow$ & CLIPScore $\uparrow$  & Aesthetics $\uparrow$ & GLCM Score $\uparrow$ & Compression Ratio $\downarrow$ \\ 
    \midrule
    \multirow{2}{*}{Flux-F16-WLF@2048} & Aesthetic-Train~\cite{zhang2025diffusion4k} & 39.49 & 34.41 & 6.37 & 0.61 & 13.60 \\
    & Aesthetic-Train-V2 & 39.05 & 34.34 & 6.36 & 0.67 & 12.61 \\  
    \midrule
    \multirow{2}{*}{Flux-F16-SC-WLF@2048} & Aesthetic-Train~\cite{zhang2025diffusion4k} & 38.46 & 34.38 & 6.37 & 0.71 & 10.62 \\
    & Aesthetic-Train-V2 & 38.38 & 34.42 & 6.37 & 0.79 & 9.95 \\  
    \bottomrule
  \end{tabular}
\end{table*}

\begin{figure*}
\begin{center}
\subfloat[Flux-F16-SC-WLF@2048 fine-tuned on Aesthetic-Train.]{
\centering
\includegraphics[width=\linewidth]{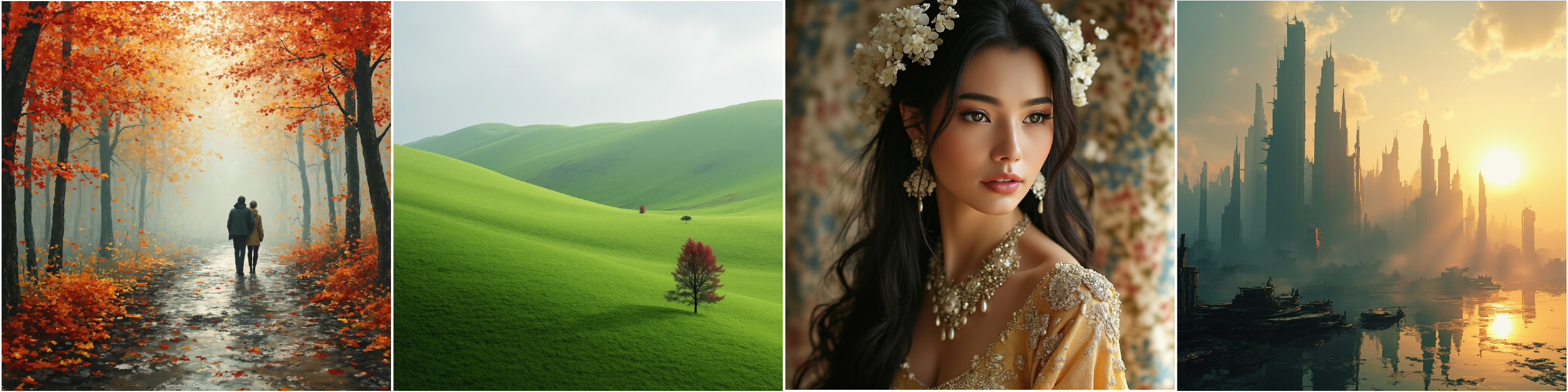}
\label{fig:scalability_v1}
}
\hfill
\subfloat[Flux-F16-SC-WLF@2048 fine-tuned on Aesthetic-Train-V2.]{
\centering
\includegraphics[width=\linewidth]{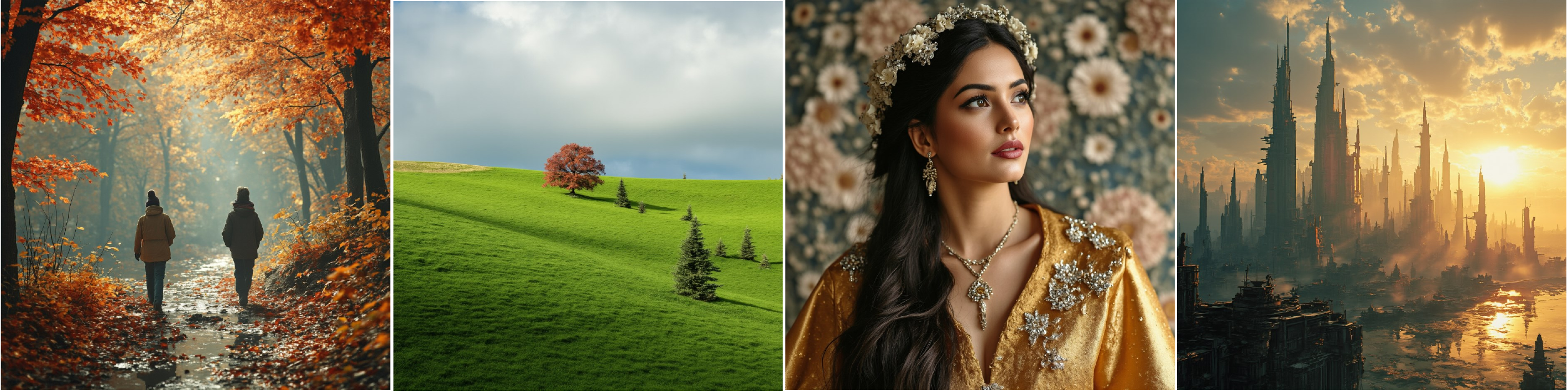}
\label{fig:scalability_v2}
}
\caption{Qualitative scalability results on Aesthetic-Eval@2048 at $2048 \times 2048$ resolution.  } 
\label{fig:scalability}
\end{center}
\end{figure*}

We conduct scalability experiments with the state-of-the-art latent diffusion model, Flux-12B, trained on both Aesthetic-Train~\cite{zhang2025diffusion4k} and Aesthetic-Train-V2 for 20K and 50K steps, respectively, and provide both quantitative and qualitative comparisons for scalability analysis. 
As shown in \cref{tab:ablation_scaling_data}, the holistic metrics, such as FID~\cite{heusel2017gans}, Aesthetics~\cite{schuhmann2022laion}, and CLIPScore~\cite{hessel2021clipscore}, tend to saturate as the volume of training data increases.
In contrast, the proposed local metrics, including the GLCM Score and Compression Ratio, demonstrate consistent and substantial improvements with the expansion of the training dataset. 
In addition to the quantitative evaluation, as illustrated in \cref{fig:scalability}, we present the generated images from different latent diffusion models for qualitative comparisons.
Notably, the images in \cref{fig:scalability_v2}, generated by the Flux-F16-SC-WLF model trained on Aesthetic-Train-V2 with a larger training set, exhibit richer textures and finer details compared to those in \cref{fig:scalability_v1}, which were generated by the model trained on Aesthetic-Train~\cite{zhang2025diffusion4k}.

Overall, both quantitative and qualitative results highlight the benefits of scalable training data in improving fine details.
Furthermore, the experimental findings show the limitations of conventional holistic metrics in evaluating ultra-high-resolution image synthesis and emphasize the necessity and effectiveness of incorporating local metrics such as the GLCM Score and Compression Ratio as supplementary indicators for assessing rich textures and fine details, thereby enabling a more comprehensive evaluation.

\subsection{Ablation Studies}

\begin{table}
  \centering
  \caption{Quantitative reconstruction results of SC-VAE with scale consistency on Aesthetic-Train at $2048 \times 2048$ resolution. SD3-VAE-F16 and Flux-VAE-F16 represent the partitioned VAE without finetuning decoder. SC and FT denote fine-tuning the decoder of the VAE with and without scale consistency respectively.
  }
  \label{tab:ablation_vae_reconstruction}
  \resizebox{\linewidth}{!}{
  \begin{tabular}{c|c|c|c|c|c}
    \toprule
    Model & rFID $\downarrow$ & NMSE $\downarrow$ & PSNR $\uparrow$ & SSIM $\uparrow$ & LPIPS $\downarrow$ \\
    \midrule
    SD3-VAE-F16~\cite{zhang2025diffusion4k} & 1.65 & 0.09 & 27.24 & 0.75 & 0.17 \\ 
    SD3-VAE-F16-FT & 0.95 & 0.09 & 28.39 & 0.79 & 0.10 \\ 
    SD3-VAE-F16-SC & \textbf{0.65} & \textbf{0.08} & \textbf{29.90} & \textbf{0.80} & \textbf{0.09} \\ 
    \midrule
    Flux-VAE-F16~\cite{zhang2025diffusion4k} & 1.95 & 0.10 & 27.54 & 0.77 & 0.17 \\ 
    Flux-VAE-F16-FT & 0.83 & 0.08 & 30.34 & 0.82 & 0.09 \\ 
    Flux-VAE-F16-SC & \textbf{0.55} & \textbf{0.06} & \textbf{32.01} & \textbf{0.85} & \textbf{0.07} \\ 
    \bottomrule
  \end{tabular}
  }
\end{table}

\noindent\textbf{Ablation on Scale Consistency.}
To evaluation the effectiveness and generalization of the proposed scale consistency regularization approach in the SC-VAE, we provide the quantitative reconstruction performance of different VAEs with a down-sampling factor of $F=16$ on Aesthetic-Train at $2048 \times 2048$ resolution. 
As shown in \cref{tab:ablation_vae_reconstruction}, our SC-VAE outperforms both the baseline partitioned VAE~\cite{zhang2025diffusion4k} (without fine-tuning) and the vanilla VAE fine-tuning approaches without scale consistency~\cite{esser2021taming, rombach2022high} across all evaluation metrics, including rFID, NMSE, PSNR, SSIM, and LPIPS, for both SD3-VAE and Flux-VAE. 
The quantitative results further emphasize the effectiveness of the proposed scale consistency regularization approach in reconstructing ultra-high-resolution images, ensuring latent space consistency and eliminating potential distribution shifts for subsequent fine-tuning of diffusion models.

\begin{table}
  \centering
  \caption{Ablation study of WLF on Aesthetic-Eval@4096 at $4096 \times 4096$ resolution. SD3-F16-FT@4096 represents fine-tuning the diffusion model without WLF.}
  \label{tab:ablation_wlf}
  \resizebox{\columnwidth}{!}{
  \begin{tabular}{c|c|c|c|c}
    \toprule
    Model & CLIPScore $\uparrow$ & Aesthetics $\uparrow$ & GLCM Score $\uparrow$ & Compression Ratio $\downarrow$  \\ 
    \midrule
    SD3-F16@4096 & 33.12 & 5.97 & 0.73 & 11.97 \\
    SD3-F16-FT@4096 & 34.14 & 5.99 & 0.74 & 11.41 \\ 
    SD3-F16-WLF@4096 & \textbf{34.40} & \textbf{6.07} & \textbf{0.77} & \textbf{10.50} \\ 
    \bottomrule
  \end{tabular}
  }
\end{table}

\begin{figure}
\begin{center}
\subfloat[Fine-tuning without WLF.]{
\centering
\includegraphics[width=.468\linewidth]{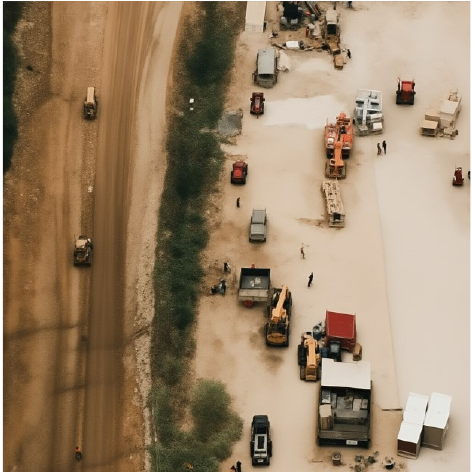}
\label{fig:ablation_without_wlf}
}
\hfill
\subfloat[Fine-tuning with WLF.]{
\centering
\includegraphics[width=.468\linewidth]{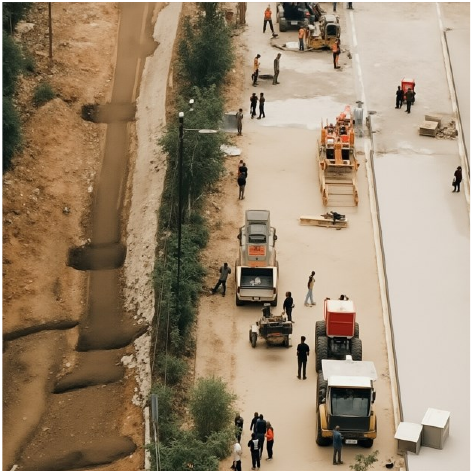}
\label{fig:ablation_with_wlf}
}
\caption{Qualitative ablation study on WLF. The image in \cref{fig:ablation_with_wlf}, generated with the WLF model, exhibit richer details than those in \cref{fig:ablation_without_wlf}.  } 
\label{fig:ablation_wlf}
\end{center}
\end{figure}

\noindent\textbf{Ablation on WLF.}
To demonstrate the effectiveness of the WLF training objective in \cref{eq:wavelet_rectified_flow}, we conduct ablation studies with SD3, comparing fine-tuning diffusion models with and without the WLF objective.
The experimental results for Aesthetic-Eval@4096 are presented in \cref{tab:ablation_wlf}.
Compared to fine-tuning without WLF, our WLF method demonstrates superior performance in CLIPScore~\cite{hessel2021clipscore}, Aesthetics~\cite{schuhmann2022laion}, GLCM Score, and Compression Ratio, significantly highlighting its effectiveness in improving visual aesthetics, prompt adherence, and high-frequency details.

In addition to the quantitative analysis, we provide qualitative comparisons of latent fine-tuning with and without WLF to further showcase its impact. 
To ensure a fair comparison, we use the same random seeds and text prompts across the experiments.
As illustrated in \cref{fig:ablation_wlf}, images generated using WLF exhibit noticeably richer details compared to those generated without WLF, clearly demonstrating the effectiveness of our method in enhancing fine details.

\begin{table}
  \centering
  \caption{Ablation study on quality of image captions on Aesthetic-Eval@4096 at $4096 \times 4096$ resolution. }
  \label{tab:ablation_captions}
  \begin{tabular}{l|c|cc}
    \toprule
    Captions & Model & CLIPScore $\uparrow$ & Aesthetics $\uparrow$  \\ 
    \midrule
    LAION-5B & SD3-F16@4096 & 29.37 & 5.90 \\ 
    GPT-4o & SD3-F16@4096 & \textbf{33.12} & \textbf{5.97} \\ 
    \midrule
    LAION-5B & Flux-F16@4096 & 29.12 & 6.02 \\ 
    GPT-4o & Flux-F16@4096 & \textbf{33.67} & \textbf{6.11} \\ 
    \bottomrule
  \end{tabular}
\end{table}

\noindent\textbf{Ablation on Quality of Image Captions.}
We compare the performance of 4K image synthesis using both original captions from LAION-5B~\cite{schuhmann2022laion} and captions generated by GPT-4o.
As shown in \cref{tab:ablation_captions}, both SD3 and Flux exhibit improved results in Aesthetic-Eval@4096 when utilizing captions generated by GPT-4o.
Quantitative results demonstrate that prompts generated by GPT-4o significantly enhance image synthesis quality and prompt coherence, underscoring the critical role of high-quality prompts in 4K image generation and the effectiveness of captions generated by GPT-4o in Aesthetic-Eval.

\section{Conclusion}

In this paper, we present Diffusion-4K, a novel framework for direct ultra-high-resolution image synthesis utilizing text-to-image diffusion models. 
We introduce the Aesthetic-4K benchmark to address the lack of a publicly available 4K image synthesis dataset and propose comprehensive assessments for ultra-high-resolution image generation.
Additionally, we design the scale consistent VAE and wavelet-based latent fine-tuning, capable of training with state-of-the-art latent diffusion models at $4096 \times 4096$ resolution, such as SD3 and Flux.
Both qualitative and quantitative results demonstrate the effectiveness and generalization of our approach in training and generating photorealistic 4K images, particularly in visual aesthetics, prompt adherence, and fine details.

However, our approach is not without limitations. 
Our method fine-tunes the base diffusion models and, as such, inherit their limitations, potentially lacking the ability to generate certain specific scenes and objects.



\bibliographystyle{IEEEtran}
\bibliography{reference.bib}

\end{document}